\documentclass[12pt, a4paper, twoside]{article}

\usepackage[utf8]{inputenc}
\usepackage[T1]{fontenc}
\usepackage{amsmath}
\usepackage{amssymb}
\usepackage{graphicx}
\usepackage{xcolor}
\usepackage{url}
\usepackage{orcidlink}
\usepackage{subcaption}
\usepackage{hhline}
\usepackage{wrapfig}
\usepackage{multirow}
\usepackage{booktabs}
\usepackage{array}
\usepackage{natbib}
\usepackage{hyperref}
\hypersetup{colorlinks=true,linkcolor=blue,urlcolor=cyan,citecolor=blue}
\usepackage[margin=1in]{geometry}
\title{No Free Lunch for Synthetic Images under Data Scarcity Conditions}

\author{
  Borja Arroyo Galende$^1$\orcidlink{https://orcid.org/0000-0001-5035-0998} \and
  Alejandro Almodóvar$^1$\orcidlink{https://orcid.org/0009-0006-0900-4026} \and
  Patricia A. Apellániz$^1$\orcidlink{https://orcid.org/0000-0002-8604-9758} \and
  Juan Parras$^1$\orcidlink{https://orcid.org/0000-0002-7028-3179} \and
  Silvia Uribe$^2$\orcidlink{https://orcid.org/0000-0001-6156-5492} \and
  Santiago Zazo$^1$\orcidlink{https://orcid.org/0000-0001-9073-7927}
}

\date{\today}

\newcommand{\aff}[1]{$^{#1}$}

\begin{document}

\maketitle

\begin{center}
\begin{small}
\begin{tabular}{p{\textwidth}}
\hline
\aff{1}Information Processing and Telecommunications Center, Escuela Técnica Superior de Ingeniería de Telecomunicación, Universidad Politécnica de Madrid, Madrid, 28030, Spain \\
\aff{2}Escuela Técnica Superior de Ingeniería de Sistemas Informáticos, Universidad Politécnica de Madrid, Madrid, 28030, Spain\\ \hline
\end{tabular}
\end{small}
\end{center}

\vspace{1em}

\begin{abstract}
This study investigates the trade-offs between fidelity, privacy, and utility in synthetic data generation under conditions of data scarcity and privacy sensitivity. We propose an evaluation framework that jointly assesses these three dimensions and apply it to three widely used generative models, VAE, GAN, and DDPM. The evaluation spans three image datasets, MNIST, OCTMNIST, and OrganAMNIST, encompassing both general-purpose and medical imaging domains. Notable differences arise between the three models in their behaviour when differential privacy mechanisms are introduced during training. GAN and DDPM demonstrate greater robustness, maintaining higher fidelity and downstream utility across a range of noise levels, while VAE degrades more rapidly as privacy constraints increase. This study highlights the importance of a multidimensional evaluation of deep generative models, also noting that their behaviour significantly differs when privacy techniques are applied.
\end{abstract}

**Keywords:** machine learning, deep generative model, synthetic data, evaluation, trade-off


\section{Introduction}

Synthetic data generation has emerged as a critical tool in data science and machine learning, offering solutions to significant practical challenges, including data scarcity, privacy concerns, and limitations arising from sensitive or regulated information \cite{rashidi_novel_2024, mendes_synthetic_2025, damico_synthetic_2023}. With the advent of sophisticated generative models such as Generative Adversarial Networks (GAN) \cite{goodfellow_generative_2020}, Variational Autoencoders (VAE) \cite{kingma_auto-encoding_2014}, and Denoising Diffusion Probabilistic Models (DDPM) \cite{ho_denoising_2020}, it has become possible to synthesise realistic data across diverse domains, particularly in high-dimensional modalities such as images \cite{eraslan_deep_2019, creswell_generative_2018}.

Images represent structured and intuitive representations of information that align closely with human perception. Thus, much of the foundational work in modern generative modelling has drawn inspiration from our understanding of image processing and representation learning \cite{lecun_convolutional_1998}. Despite these advances, image data synthesis remains inherently challenging due to the complexity and high dimensionality of image data \cite{tsirikoglou_survey_2020}. This complexity exacerbates the problems associated with the curse of dimensionality \cite{bellman_mathematical_1959}, requiring models to incorporate a large number of degrees of freedom to effectively capture the underlying data distribution \cite{bengio_representation_2013}.

Generative models operate by learning parameterised representations of data distributions \cite{bengio_representation_2013}. Formally, they aim to approximate the probability distribution $p(x)$ of the observed data $x$ by introducing latent variables $z$ and modelling a conditional distribution $p_\theta(x \mid z)$, where $\theta$ encapsulates the model parameters \cite{ruthotto_introduction_2021}. For image datasets, the underlying distributions typically exhibit complex and multimodal characteristics, which require highly flexible architectures with extensive parameterisation \cite{bengio_representation_2013, eraslan_deep_2019, forsyth_computer_2002}. However, as the complexity and flexibility of these models increase, they often demand large volumes of training data to achieve satisfactory performance \cite{goodfellow_deep_2016}.

In privacy-sensitive applications, such as healthcare, finance, and biometric systems, directly using or sharing real data sets can pose serious ethical, legal, and regulatory risks. For the latter, extensive guidance is provided in the European Health Data Space Regulation (EHDS), Regulation (EU) 2025/327 \cite{directorate-general_for_health_and_food_safety_dg_sante_european_commission_regulation_2025}, and in the updated 2025 version of the HIPAA Privacy Rule \cite{us_department_of_health_and_human_services_office_for_civil_rights_notice_2025}. Synthetic data generation offers an attractive alternative by enabling the creation of realistic yet artificial data sets that preserve essential statistical properties while mitigating privacy concerns \cite{mendes_synthetic_2025, giuffre_harnessing_2023, gonzales_synthetic_2023}. However, synthesising data with privacy guarantees introduces inherent trade-offs: enhancing privacy often requires introducing noise or constraints that may degrade the fidelity and downstream utility of the data \cite{adams_fidelity_2025, kaabachi_scoping_2025}.

Therefore, validating synthetic data in privacy-aware contexts requires a comprehensive assessment of three fundamental dimensions: fidelity, privacy, and utility \cite{jordon_synthetic_2024}. Fidelity refers to the similarity between the synthetic and real data distributions, which is used to evaluate the accuracy with which synthetic data replicate the statistical characteristics of the original data. Utility assesses the usefulness of synthetic data for performing downstream tasks, such as classification or regression, thus measuring the practical value of synthetic datasets. Privacy evaluates the degree to which synthetic data protect against the disclosure of sensitive information, typically quantified through differential privacy measures and leakage analysis.

This article examines the fundamental trade‑off between fidelity, privacy, and utility under data scarcity. We empirically show that improving one of these dimensions necessarily compromises at least one of the others. This phenomenon aligns with the no-free lunch theorem for trade-offs between privacy and utility, which formally states that improving privacy guarantees incurs a bounded degradation in utility, and vice versa \cite{zhang_no_2022, kifer_no_2011}. Using a unified evaluation framework, we compare GAN, DDPM, and VAE in fidelity, privacy, and utility. Our results provide clear guidance on the use of generative models in privacy‑sensitive settings.

\section{Objectives and contributions}
This study proposes a unified empirical framework for systematically evaluating synthetic image generation under conditions of data scarcity and differential privacy. Unlike previous frameworks predominantly designed for tabular data, such as SynthEval \cite{lautrup_syntheval_2024} and multidimensional benchmarking by Sidorenko et al. \cite{sidorenko_benchmarking_2025}, our methodology focusses on high-dimensional data from the image modality. Hence, the main contributions of this study are as follows.

\begin{itemize}
    \item Introduce and validate a unified evaluation framework that jointly quantifies the fidelity, privacy, and utility of synthetic images under data scarcity constraints.

    \item Apply this framework to conduct, to the best of our knowledge, the first empirical comparison of deep generative architectures across these three dimensions for image datasets, revealing distinct performance patterns and trade-offs under varying levels of privacy.

    \item Demonstrate the novelty and practical relevance of evaluating synthetic data generation in medical imaging scenarios, providing evidence-based insights for model selection in domains subject to regulatory requirements such as EHDS and HIPAA.

    \item All the code related to the experiments run as part of this article is hosted at \url{https://github.com/BorjaArroyo/synthetic-images-tradeoff}. In this way, our method can be replicated in other use cases.
\end{itemize}

\section{Methods}
This study adopts a multidimensional evaluation framework to assess the quality of synthetic data produced by generative models. Specifically, three core dimensions are considered: privacy, fidelity, and utility. These dimensions capture complementary aspects of synthetic data quality: its robustness to privacy risks, its similarity to real data, and its performance in downstream tasks. By analysing these axes jointly, we provide a comprehensive characterisation of model behaviour under differentially private training settings.

The generative models evaluated in this work include VAE, GAN, and DDPM, which are among the most widely used architectures for the synthesis of high-dimensional data \cite{lu_machine_2024}. The following subsections describe the modelling assumptions and experimental setup used to evaluate each dimension.

\subsection{Generative Modelling}
Generative modelling seeks to approximate the data distribution \( p(x) \) using data-driven mechanisms \cite{zhu_generative_2024}. In this study, we consider three prominent generative architectures. These models approach data synthesis through fundamentally different principles.

\begin{itemize}
    \item VAE models the data distribution explicitly. They define a likelihood function \( p_\theta(x \mid z) \) conditioned on a latent variable \( z \sim p(z) \), and approximate the intractable posterior \( p(z \mid x) \) with a variational distribution \( q_\phi(z \mid x) \). The model is trained to maximise the evidence lower bound (ELBO), balancing reconstruction quality and regularisation. This probabilistic framework enables VAEs to produce diverse samples and estimate data likelihoods, although they often suffer from over-smoothing \cite{takida_preventing_2022}.

    \item GAN adopts a game-theory training approach rather than an explicit probabilistic model. A generator \( G(z) \), which maps random noise \( z \sim p(z) \) to synthetic data, competes against a discriminator \( D(x) \), which tries to distinguish real from generated samples. The generator is optimised to produce samples that the discriminator cannot reliably differentiate from real data. GANs are known to generate high-fidelity images \cite{lv_which_2021}, but lack an explicit likelihood function and can suffer from mode collapse or training instability \cite{goodfellow_generative_2020}.

    \item DDPM define a generative process through a sequence of denoising steps, learning to reverse a gradual corruption of data with noise. Starting from pure noise $p(z) \equiv x_T \sim \mathcal{N}(0, I)$, the model learns to recover the original data distribution $p(x)$ by estimating the reverse transitions $p_\theta(x_{t-1} \mid x_t)$. A neural network is trained to predict either the added noise or the original data sample at each step. This iterative framework offers strong sample diversity and state-of-the-art image quality and allows for explicit log-likelihood estimation under certain configurations. However, their slow sampling due to numerous denoising steps can limit their applicability \cite{ho_denoising_2020}.
\end{itemize}

In all architectures, a differentially private approach known as DPSGD (differentially private stochastic gradient descent) \cite{song_stochastic_2013} is used to provide output privacy to the generative model in the experiments. Given a normal distribution $d \sim N(0, \sigma^2)$, the backward pass is characterised by adding a perturbation to the clipped gradients ($g$) such that $g^* = g + d$. Therefore, for $\sigma = 0$, we can directly apply stochastic gradient descent (SGD). The way in which DPSGD is applied to each of the models differs, as in VAE and DDPM, the entire architecture is typically affected by noise disturbance, as in \cite{takahashi_differentially_2020, dockhorn_differentially_2023}, while for GAN, only the discriminator upgrades are typically noised, as in \cite{abadi_deep_2016}.

In addition, conditional versions of the three models were used to facilitate the sampling process. Conditional sampling takes a condition $y$ to produce a sample. If seen as a distribution, it represents $p(x|y=y_k)$, where $y_k$ is a label extracted from the data. One of the key elements of the synthetic data quality evaluation is class balance to avoid bias. Using conditional sampling, the models can be forced to generate an underrepresented class, which is impossible for unconditional models.

\subsection{Privacy Evaluation}
Overfitting is a well-known risk in flexible models with numerous degrees of freedom \cite{roelofs_meta-analysis_2019, bejani_systematic_2021}. In generative settings, it rarely manifests itself as an exact replication of training data but instead through a localised concentration of probability mass around the seen samples \cite{loaiza_ganem_diagnosing_2022}. Such distortions of the learnt distribution are often imperceptible to humans but raise important concerns about generalisation and privacy \cite{carlini_extracting_2021}. This is particularly relevant when models are trained on sensitive data and synthetic outputs are publicly released.

DP provides a framework that quantifies the notion of overfitting in a rigorous manner. Recent work has proposed a formalisation of DP specifically tailored to generative models \cite{galende_membership_2025}. In the remainder of this section, we present the methodology used to evaluate privacy from a DP perspective.

Let $X_t$ and $X_t'$ be two neighbouring data sets that differ only on a sample referred to as query $x_q$, that is, $X_t = X_t' \cup \{x_q\}$. Let $\theta_v$ and $\theta_a$ be the parameters of two identical generative models trained on the data sets $X_t$ and $X_t'$. The \emph{victim} model $\theta_v$ is trained on $X_t$, and the \emph{ablated} model $\theta_a$ is trained on $X_t'$, that is, a data set that excludes the query sample $x_q$. Privacy leakage is quantified by the following expression:

\begin{equation} \label{eq:dp-epsilon-sup}
    \varepsilon(\theta_v, \theta_a)
    = \sup_{x_q \in \mathcal{X}}
    \left[
        \ln \frac{
            p(x_q \mid \theta_v)
        }{
            p(x_q \mid \theta_a)
        }
    \right]
\end{equation}

In practice, this supremum is approximated by selecting $x_q^*$ from a minority class or underrepresented region in the data space, yielding:

\begin{equation} \label{eq:dp-epsilon-estimate}
    \varepsilon(\theta_v, \theta_a)
    \approx
    \ln \frac{
        p(x_q^* \mid \theta_v)
    }{
        p(x_q^* \mid \theta_a)
    }
\end{equation}

This formulation highlights how local changes in the likelihood of the model reflect privacy vulnerabilities. Moreover, this idea is naturally connected with the curse of dimensionality \cite{bellman_mathematical_1959}, where overfitting in high-dimensional spaces can result in disproportionately sharp density peaks.

\begin{figure}[!ht]
    \centering
    \includegraphics[width=0.75\linewidth]{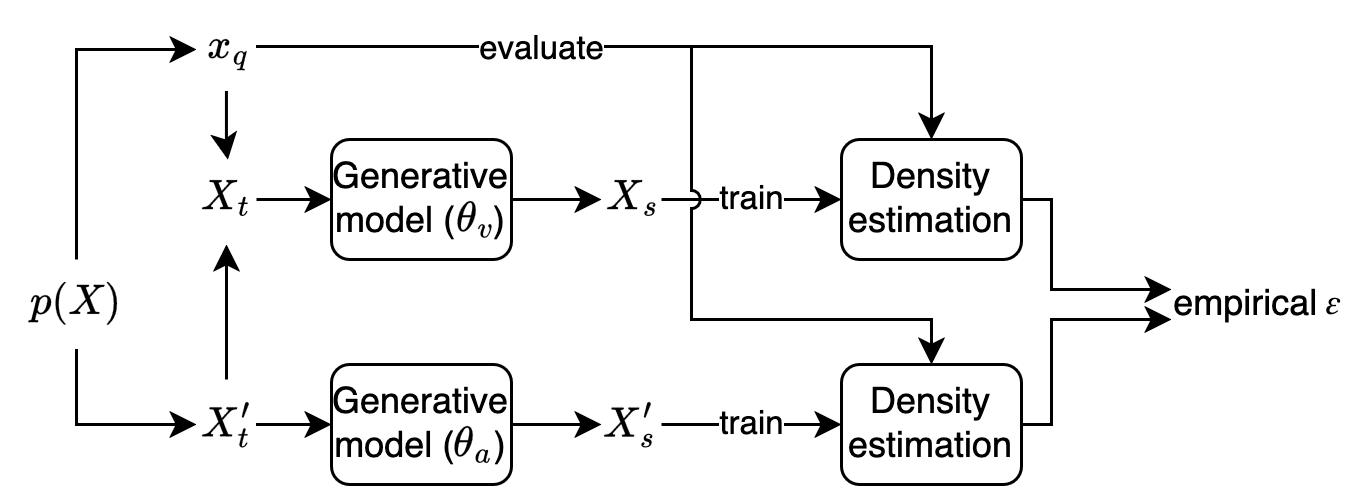}
    \caption{Privacy estimation framework. $X$ is the original dataset; $p(X)$ its distribution; $x_q$ is the query sample excluded in the ablated model; $X_s$ and $X_s'$ are synthetic datasets generated by the victim ($\theta_v$) and ablated ($\theta_v$) models, respectively. A large sample size is used to estimate $p(x_q^*)$ under both models, enabling calculation of $\varepsilon$ by using Equation \eqref{eq:dp-epsilon-estimate}.}
    \label{fig:flow-diagram-privacy}
\end{figure}

Our empirical estimation of $\varepsilon$ follows the procedure of \cite{galende_membership_2025}. First, we train two generative models on neighbouring datasets as defined above. The victim model has access to the query’s class, while the ablated model does not. Each model then generates a large synthetic data set with balanced class distributions. Then, two similar density estimation models are trained on the synthetic sets, one for the ablated set and another for the victim set. The densities at $x_q^*$ are evaluated under both models and used to compute Equation \eqref{eq:dp-epsilon-estimate}. Figure \ref{fig:flow-diagram-privacy} outlines this process. Section \ref{sec:limitations} provides a discussion of the complexity of density estimation in high-dimensional and data-scarce settings.

In addition to this metric, the empirical cumulative distribution functions (ECDF) of the minimum distances from the synthetic and holdout samples to the training are compared. Synthetic samples that are too close to real training data may indicate memorisation. When the ECDF of synthetic distances increases more rapidly than that of holdout distances, this suggests a higher risk of privacy. To quantify the difference between ECDF across all sigmas, for each $\sigma$ value, two metrics are computed: the area between both ECDF curves and the Kolmogorov-Smirnov (KS) distance. While the former is clear, the latter just reflects the maximum distance between both ECDF curves. Note that in both cases, we maintain the absolute value so that this magnitude can serve to reflect the balance between fidelity and privacy.

\subsection{Fidelity Evaluation}

Fidelity evaluates how closely synthetic data resemble real data in terms of statistical properties. In the context of generative models for image data, this dimension is critical, as it reflects the generator's capacity to reproduce the underlying data distribution without direct replication. Hence, synthetic image fidelity evaluation is linked to the visual properties of the image.

We apply a suite of well-established image similarity metrics to quantify fidelity. Among the most widely used is the Frechet Inception Distance (FID), which computes the Wasserstein-2 distance between multivariate Gaussians fitted to feature representations of real and synthetic images extracted by a pre-trained Inception network. Lower FID scores indicate a closer match in both mean and covariance of the feature space, suggesting greater visual and distributional similarity. Additionally, we consider the Peak Signal-to-Noise Ratio (PSNR), which measures the ratio between the maximum possible signal and the noise introduced by the image differences. Higher PSNR values correspond to higher fidelity, especially when pixel-level similarity is relevant. We also employ the Learned Perceptual Image Patch Similarity (LPIPS), which measures perceptual similarity by comparing deep feature representations of image patches extracted from neural networks; lower LPIPS values indicate greater visual similarity as perceived by human observers. Furthermore, we include the Inception Score (IS), which evaluates both quality and diversity of synthetic images independently of real data, with higher values indicating better generation quality. We also utilise the Structural Similarity Index Measure (SSIM), which assesses structural similarity by considering luminance, contrast, and structure patterns, with higher values representing better preservation of image structure. This comprehensive suite of metrics can be extended with additional fidelity measures as needed for specific evaluation requirements.

\begin{figure}[!ht]
    \centering
    \includegraphics[width=0.75\linewidth]{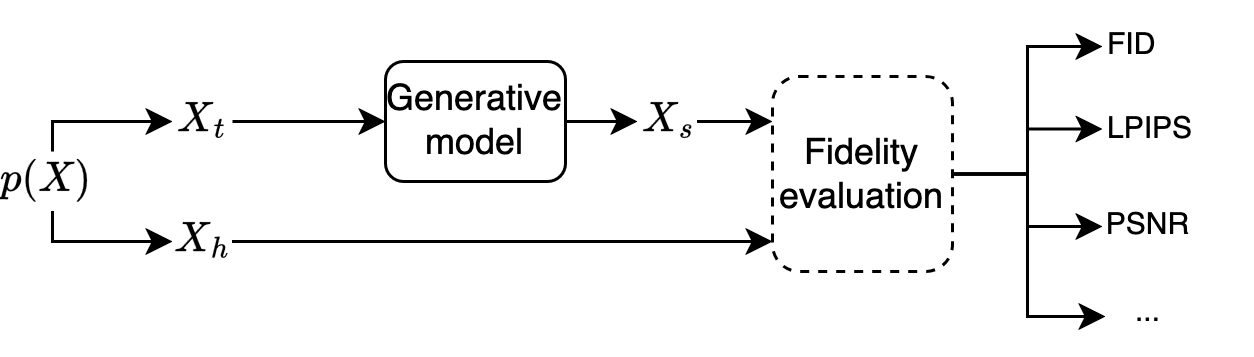}
    \caption{Design for the fidelity analysis. $X$ is the data; $p(X)$ its distribution; $X_t$ and $X_h$ are samples from $p(X)$; $X_s$ is the synthetic data set. The sizes of $X_h$ and $X_s$ are matched for fair metric estimation.}
    \label{fig:flow-diagram-fidelity}
\end{figure}

To ensure a fair comparison, the real and synthetic data sets used in metric computation are matched in sample size. Figure~\ref{fig:flow-diagram-fidelity} illustrates the data pipeline used in fidelity analysis, where $X_t$ and $X_h$ denote subsets of real data, and $X_s$ corresponds to the synthetic set sampled from the model.

In data-rich contexts, alternative fidelity measures based on bounded divergence metrics, such as the Jensen-Shannon distance, may also be employed \cite{apellaniz_synthetic_2024}. These metrics assess statistical similarity between distributions and are particularly suitable when working with latent representations or when pixel-level metrics are insufficient to capture semantic fidelity.

\subsection{Utility Evaluation}
Utility reflects the extent to which synthetic data can support downstream tasks. Even when fidelity is compromised, structural preservation may allow privacy-aware synthetic data to remain useful.

\begin{figure}[!ht]
    \centering
    \includegraphics[width=0.80\linewidth]{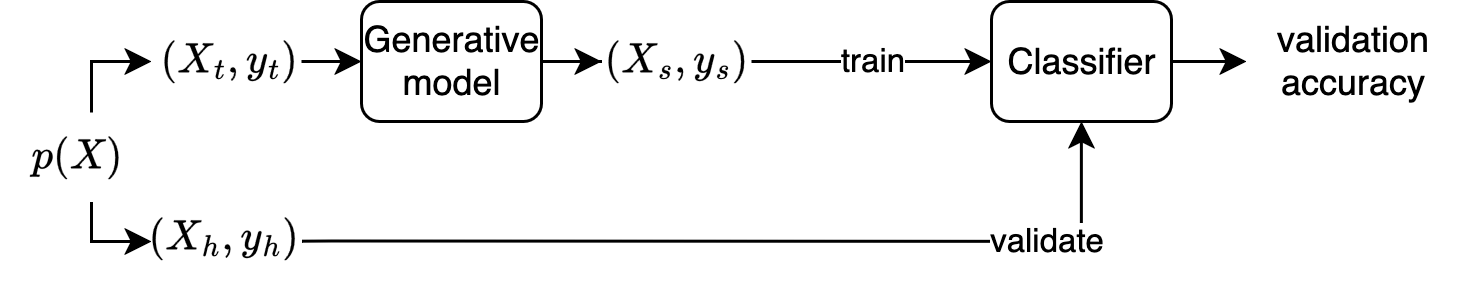}
    \caption{Design for the utility analysis. $X$ is the data; $p(X)$ its distribution; $(X_t, y_t)$ and $(X_h, y_h)$ are real labeled data; $X_s$ is synthetic. The synthetic and holdout sets have the same size to fairly assess task performance.}
    \label{fig:flow-diagram-utility}
\end{figure}

To standardise the evaluation of usefulness, all data sets in this study were selected to share a common supervised task: multiclass classification. For each dataset, a classifier is trained exclusively on synthetic data and evaluated on a held-out real test set. This design isolates the impact of synthetic data quality on downstream generalisation. Figure \ref{fig:flow-diagram-utility} demonstrates this process.

Because data scarcity significantly affects task performance, we limit our analysis to settings that enforce the same sample budget between experiments. Consequently, comparisons are made only between models trained on the same number of synthetic samples.

\section{Results}
This section first introduces the density estimation challenges that arise in the context of privacy analysis, then gathers all the evidence from the MNIST data set, and ultimately collects the results from the medical data sets, OCTMNIST and OrganAMNIST. Exhaustive analysis of synthetic data through this multidimensional approach is mandatory in senstive domains, as is the healthcare domain, we both require synthetic samples to be private and useful.

\subsection{Density Estimation Challenges} \label{sec:limitations}
Accurate density estimation is critical to evaluate privacy leakage using Equation \eqref{eq:dp-epsilon-estimate}. However, this task becomes especially challenging under the conditions of highly dimensional and limited data, which are common in privacy-sensitive generative modelling scenarios, as is the case for healthcare data. Several approaches exist to estimate densities, each with unique trade-offs.

Traditional nonparametric methods, such as kernel density estimation (KDE) \cite{terrell_variable_1992} or nearest neighbours (kNN) \cite{mack_multivariate_1979}, avoid restrictive assumptions about data distributions, but are significantly affected by the curse of dimensionality and sensitivity to hyperparameter tuning \cite{walt_variable_2017}. Binary density ratio estimation, which transforms the estimation problem into a classification task, struggles similarly in sparse, high-dimensional settings \cite{kanamori_theoretical_2010}. To mitigate these issues, traditional methods typically require a previous dimensionality reduction stage to simplify density estimation and improve reliability.

Neural approaches, such as normalising flows \cite{rezende_variational_2015}, offer an attractive alternative by providing explicit and tractable likelihoods without the need for an explicit dimensionality reduction step. However, these methods typically require substantial amounts of training data and careful architectural tuning to perform effectively \cite{bond-taylor_deep_2022}.

Given data scarcity constraints and the challenges associated with neural methods under such conditions, our analysis adopted dimensionality reduction combined with density estimation techniques. Several dimensionality reduction techniques, including manifold-based methods like Isomap \cite{balasubramanian_isomap_2002}, UMAP \cite{mcinnes_umap_2018}, and t-SNE \cite{maaten_visualizing_2008}, neural autoencoders, and normalising flows, which were considered an implicit projection method with tractable densities, were evaluated based on their ability to preserve essential structural characteristics.

\begin{figure}[!ht]
    \centering
    \includegraphics[width=0.75\linewidth]{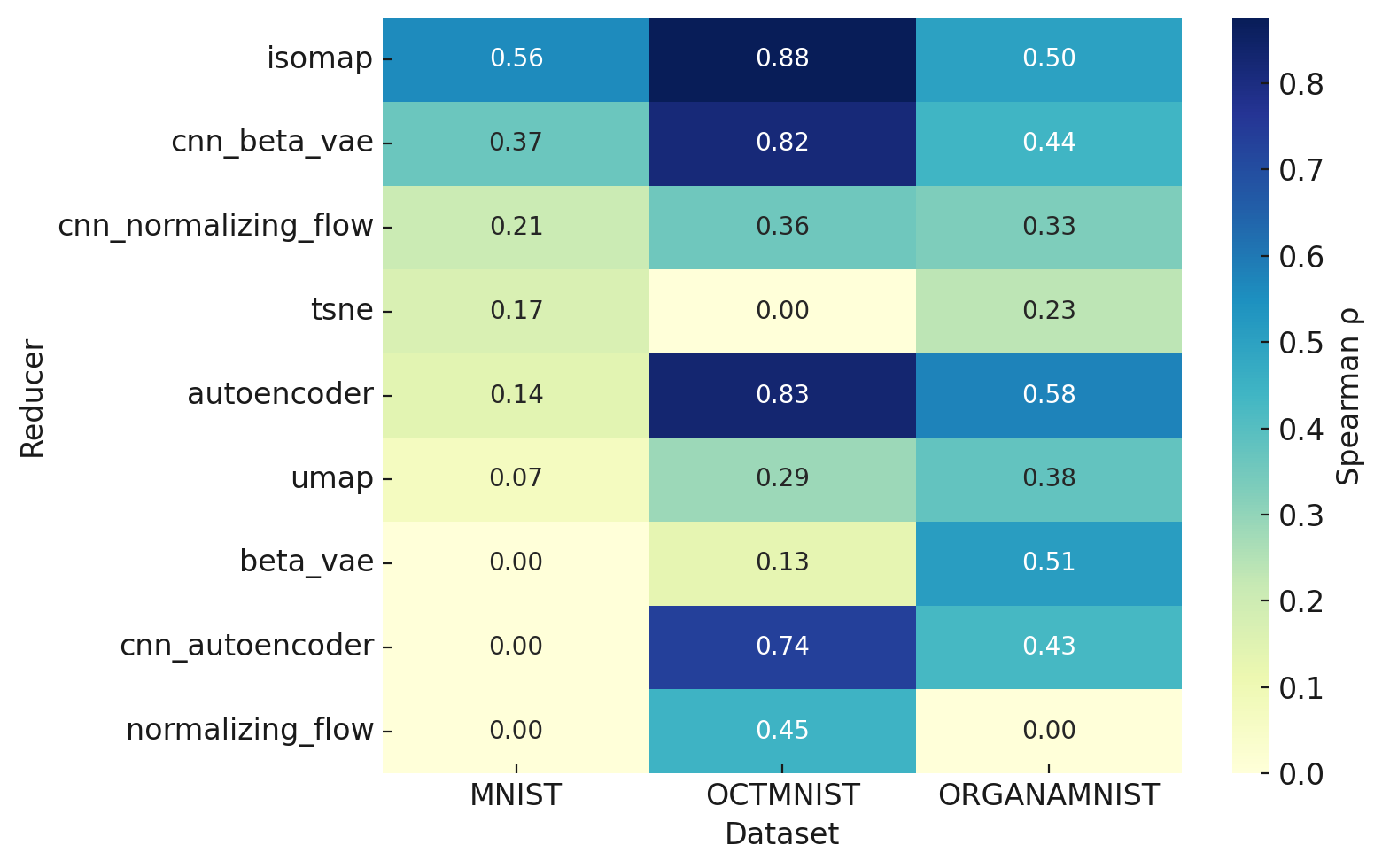}
    \caption{Spearman rank correlation coefficients across dimensionality reduction methods and datasets. Methods are sorted according to their performance on MNIST. Higher values indicate better preservation of pairwise distance rankings from the high-dimensional space to the reduced embedding; thus, higher is better. Isomap consistently outperformed other methods on the tested datasets, especially considering its robustness.}
    \label{fig:heatmap-reducers}
\end{figure}

We assessed structural preservation by comparing the ordering of pairwise distances in the original and reduced spaces using Spearman's rank correlation. This evaluation relies on the assumption that the local density at a point $x_i$ is inversely related to the distances to its nearest neighbours. Formally, this relationship can be expressed as follows:
\[
p(x_i) \propto \frac{1}{\sum_{x_j \in \mathcal{N}_k(i)} d(x_i, x_j)},
\]
where \(\mathcal{N}_k(i)\) denotes the set of the \(k\)-nearest neighbours of \(x_i\), and \(d(x_i, x_j)\) is the pairwise distance. High correlation values thus indicate effective preservation of both local and global density relationships in the embedding. Figure~\ref{fig:heatmap-reducers} summarises these evaluations. Isomap consistently provided superior performance in the datasets considered in preserving distance rankings. Consequently, Isomap was selected as the dimensionality reduction method.

In addition, several methods for estimating density ratios were compared. Of all of them, only kNN provided a consistent result in the estimates, invariant to the number of neighbours. Thus, despite kNN's known limitations in high dimensions, applying it after Isomap dimensionality reduction provided a balanced and robust approach suitable for our use case. Therefore, it was selected as the dimensionality reduction method used in the rest of the simulations in this work.

\subsection{Modelling MNIST Data}
\begin{figure}[!ht]
    \centering
    \includegraphics[width=.8\linewidth]{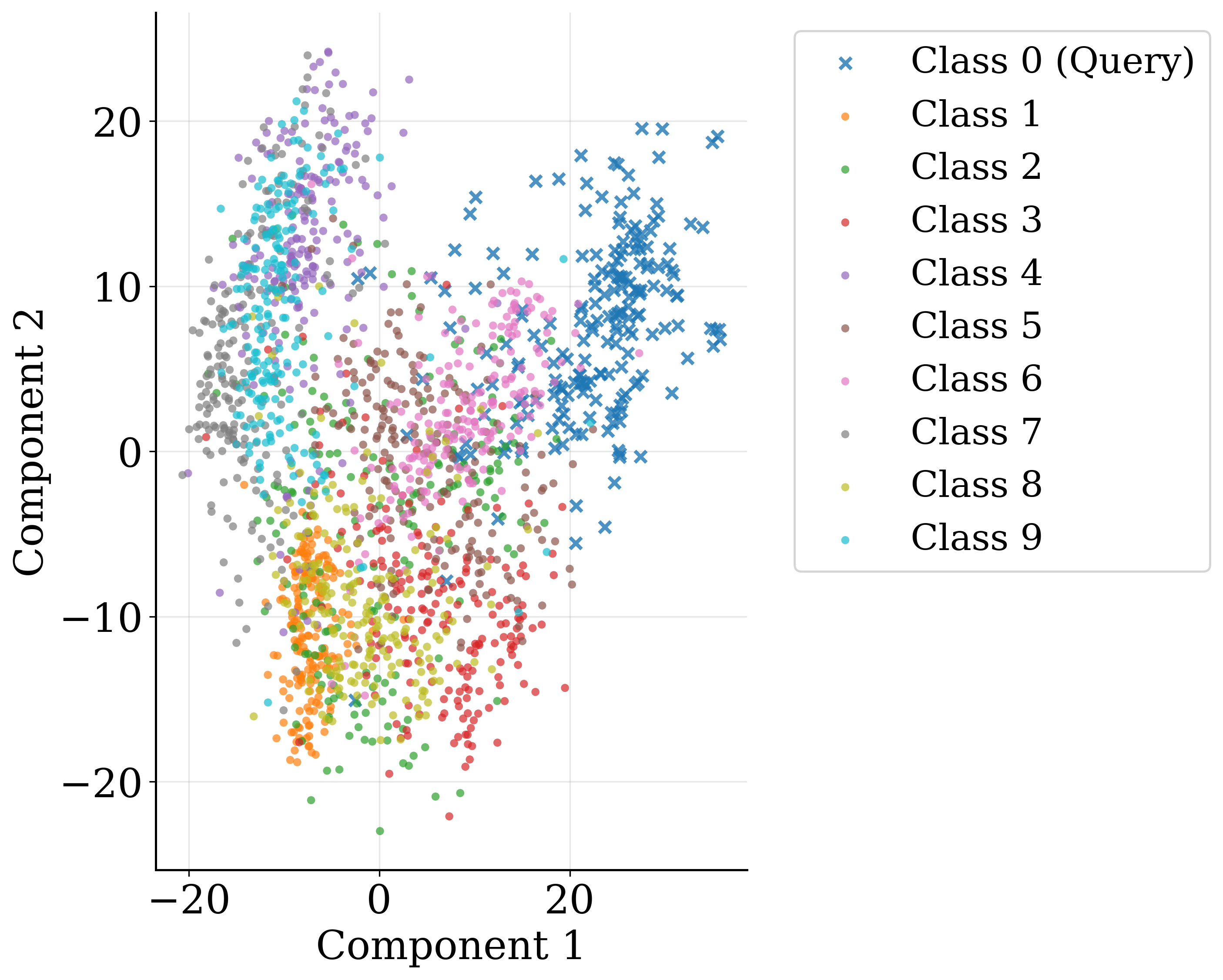}
    \caption{MNIST projected into a 2D space through the Isomap model and represented through a colour mask denoting the label of each of the samples. It is important to note that the selected class for privacy analysis is class 0, which is far from the core region and therefore potentially represents a privacy risk.}
    \label{fig:mnist-isomap-projection}
\end{figure}

This section presents a side-by-side comparison between the three architectures trained on the MNIST dataset \cite{deng_mnist_2012}. This data set has been widely used in the computer vision community and is composed of images of handwritten digits from 0 to 9 with background pixels in black and pen strokes in white. Each evaluation dimension is analysed for both models, highlighting their behaviour in terms of fidelity, privacy, and utility. For reference, Figure \ref{fig:mnist-isomap-projection} shows a scatter plot of the MNIST data set projected through the Isomap model. In this figure, we can see that the latent representation presents digits clustered on a label basis.

\subsubsection{Experimental setup}
For evaluating fidelity and utility, the model is presented with 10 samples per class, giving a total amount of 100 samples. Moreover, for privacy estimation, the selected query class is digit 0. As seen in Figure \ref{fig:mnist-isomap-projection}, this creates an overfitting risk. Hence, the victim model is fed with 10 samples per class, except for class 0, which only provides one sample, while the ablated model is trained without the 0 sample. Three seeds per setting are run, that is, the combination of model architectures and noise magnitudes, to achieve representative results.

\subsubsection{Qualitative Analysis}
\paragraph{Image Reconstruction}
Reconstruction experiments were only applicable to the VAE model as, in the case of DDPM, the reverse process is stochastic. In this setup, the encoder compresses an input image into a latent vector representation, which the decoder then reconstructs back into the image domain. Ideally, the reconstructed images would closely resemble the originals if the latent space had captured sufficient information.

\begin{figure}[!ht]
  \centering
  \begin{subfigure}[t]{0.6\textwidth}
    \centering
    \includegraphics[width=\linewidth]{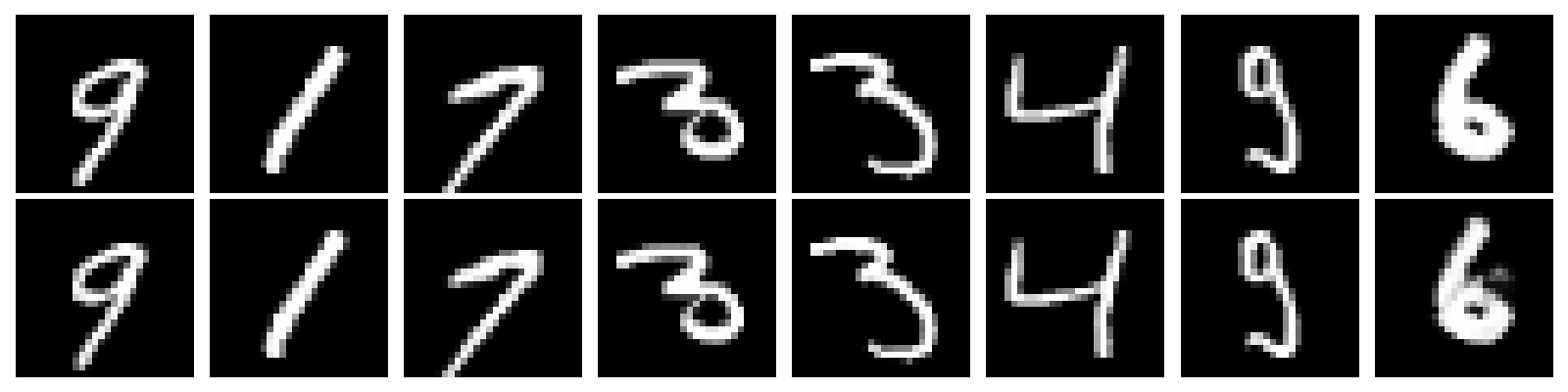}
    \caption{VAE Reconstruction with $\sigma = 0.0$}
    \label{fig:vae-recon-subfigs-a}
  \end{subfigure}
  \vspace{1em}
  \begin{subfigure}[t]{0.6\textwidth}
    \centering
    \includegraphics[width=\linewidth]{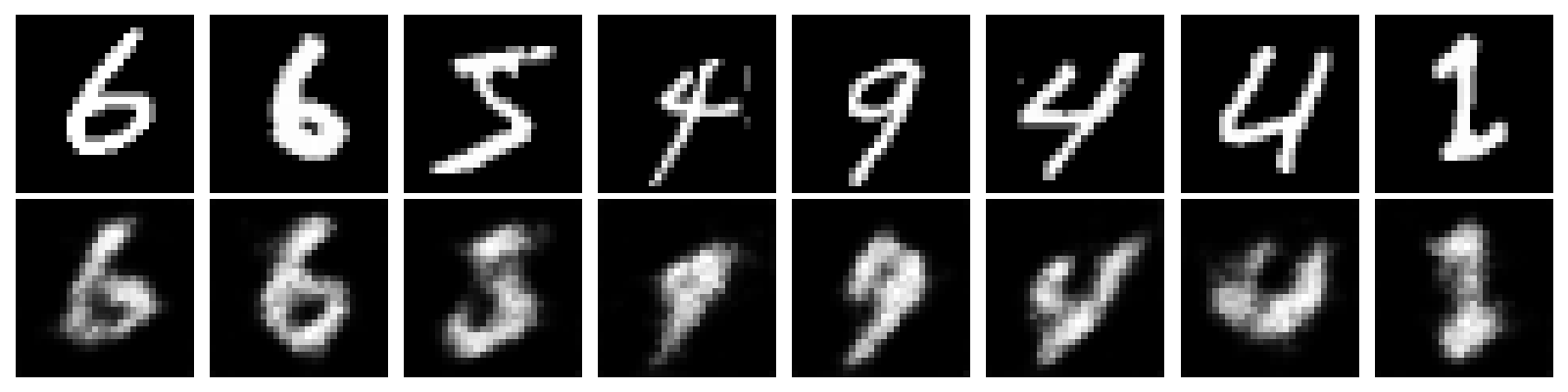}
    \caption{VAE Reconstruction with $\sigma = 0.4$}
    \label{fig:vae-recon-subfigs-b}
  \end{subfigure}
  \caption{VAE image reconstruction at different DPSGD noise levels. The upper row reflects the real images while the lower row their reconstructions. At zero noise (a), the reconstructions retain structure and clarity. At high noise (b), reconstructions are blurry and structurally altered. Note how the inclusion of privacy in the generative model has a strong impact on the visual perception of the generated digits.}
  \label{fig:vae-recon-subfigs}
\end{figure}

In the case of no perturbation ($\sigma = 0.0$), the VAE accurately reconstructed the input digits. This was expected, as the training process memorised the limited examples available, and the latent projection retained fine-grained details. Notably, as can be seen in Figure \ref{fig:vae-recon-subfigs-a}, all digits were clearly distinguishable, with only minor deviations, for example, a slight distortion in the rightmost digit.

In contrast, when noise was introduced during training (e.g. $\sigma = 0.4$), the quality of the reconstruction deteriorated substantially. Reconstructed images became blurry, and key digit features were often deformed or lost. This indicates that gradient perturbation reduced the model's capacity to memorise and reproduce training data, also affecting the sharpness and other structural properties of reconstructed data. Figure \ref{fig:vae-recon-subfigs-b} highlights these effects, where the identity of the digits becomes harder to infer and the spatial coherence degrades.

\begin{figure}[!ht]
  \centering
  \subcaptionbox{VAE \\ $\sigma=0.00$\label{fig:gen-samples-subfigs-a}}[0.15\linewidth]{
    \includegraphics[width=\linewidth]{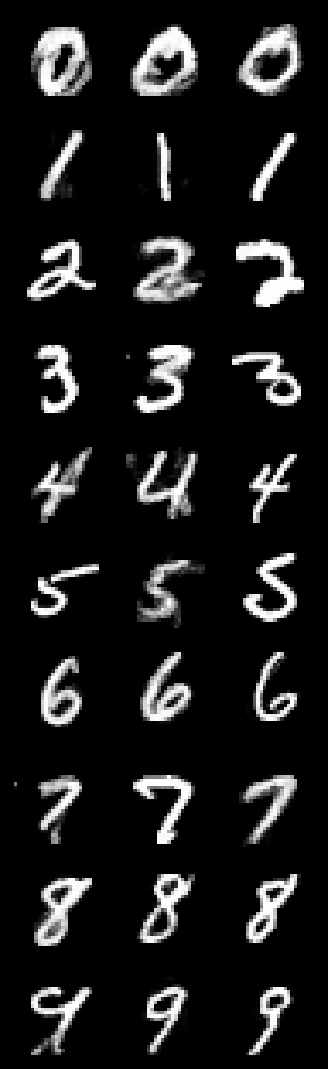}
  }
  \hspace{0.5em}
  \subcaptionbox{VAE \\ $\sigma=0.10$\label{fig:gen-samples-subfigs-b}}[0.15\linewidth]{
    \includegraphics[width=\linewidth]{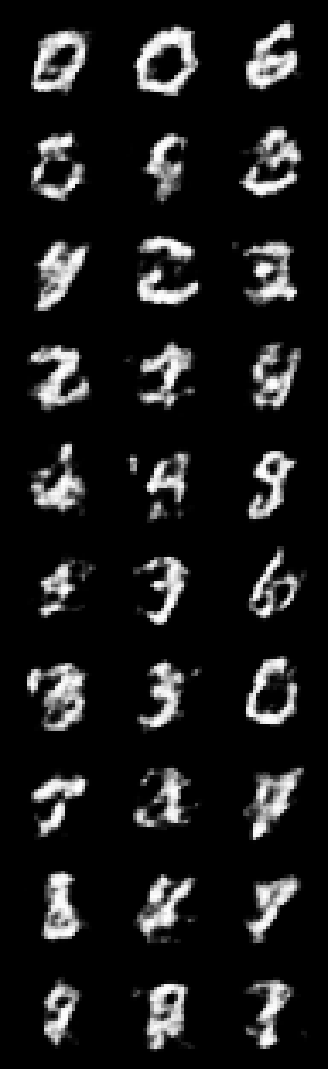}
  }
  \hspace{0.5em}
  \subcaptionbox{GAN \\ $\sigma=0.00$\label{fig:gen-samples-subfigs-c}}[0.15\linewidth]{
    \includegraphics[width=\linewidth]{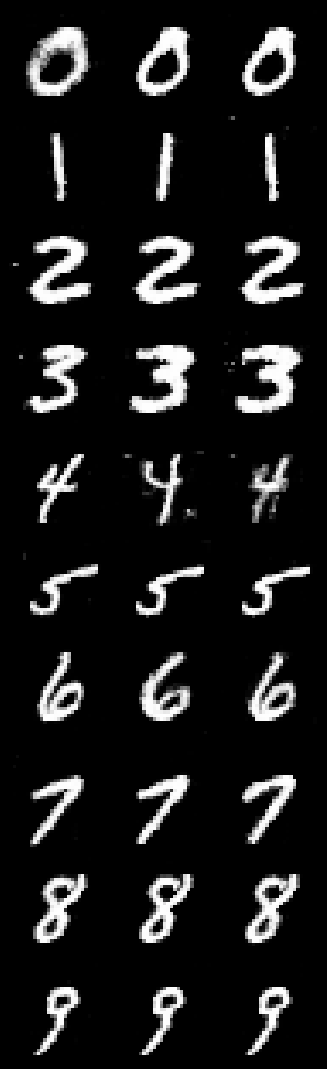}
  }

  \vspace{0.5em}
  \subcaptionbox{GAN \\ $\sigma=0.40$\label{fig:gen-samples-subfigs-d}}[0.15\linewidth]{
    \includegraphics[width=\linewidth]{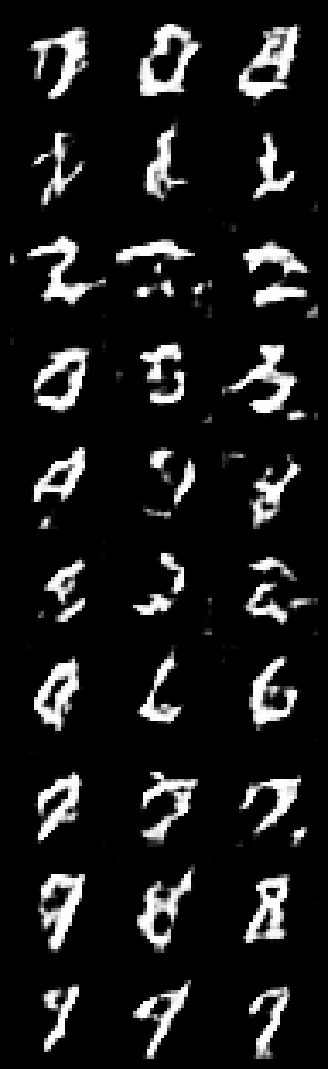}
  }
  \hspace{0.5em}
  \subcaptionbox{DDPM \\ $\sigma=0.00$\label{fig:gen-samples-subfigs-e}}[0.15\linewidth]{
    \includegraphics[width=\linewidth]{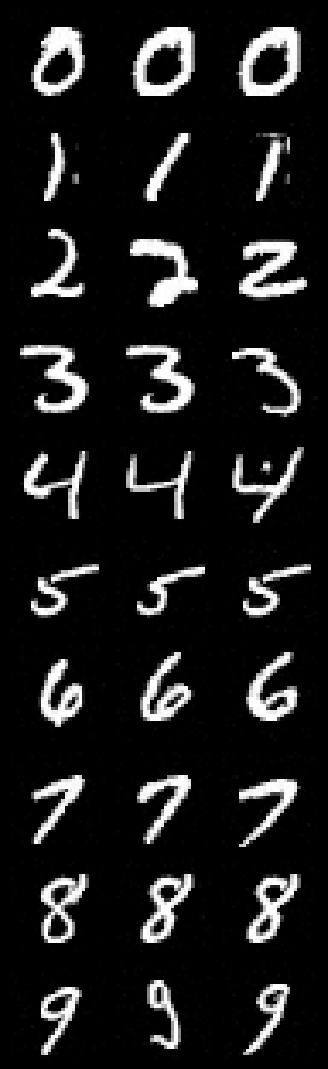}
  }
  \hspace{0.5em}
  \subcaptionbox{DDPM \\ $\sigma=0.20$\label{fig:gen-samples-subfigs-f}}[0.15\linewidth]{
    \includegraphics[width=\linewidth]{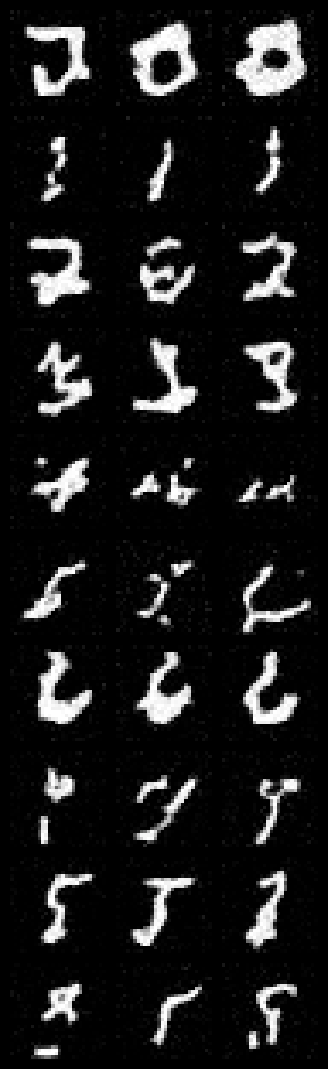}
  }

  \caption{Generated samples from all models under different DPSGD noise levels. The images are arranged with three samples per class. At $\sigma=0.00$, all models show high visual fidelity. However, VAE samples rapidly degrade, DDPM offers more robustness, tolerating higher noise values, and GAN retains spatial coherence even at very high noise values.}
  \label{fig:gen-samples-subfigs}
\end{figure}

\paragraph{Image Synthesis}
The three models were evaluated for their ability to generate synthetic samples at different $\varepsilon$-DP levels. Generation involved selecting a condition (a class label), sampling from the prior distribution over the latent space, and feeding both the latent sample and the condition to the generative mechanism. In VAE, the decoder played this role; in GAN, it was the generator; in the DDPM, it was a UNET \cite{ronneberger_u-net_2015} during the reverse process. Figure \ref{fig:gen-samples-subfigs} shows the results. Note that the discriminator in the GAN architecture was only active during training and was not used during the sampling stage.

At $\sigma=0.00$, all models performed well in generating samples, although the GAN suffered from severe mode collapse, note the low variability of the generated samples in Figure \ref{fig:gen-samples-subfigs-c}. However, even small amounts of noise significantly impacted generation quality, particularly for the VAE. As shown in Figure \ref{fig:gen-samples-subfigs-b}, the VAE output becomes blurry and more homogeneous at $\sigma = 0.10$. By comparison, the GAN at $\sigma = 0.40$ and the DDPM at $\sigma=0.20$ (Figures \ref{fig:gen-samples-subfigs-d}, \ref{fig:gen-samples-subfigs-f}) still produce samples that, although degraded, retain a certain spatial coherence and shape fidelity. This indicates that GAN and DDPM are more resistant to the effects of DPSGD noise during training. Moreover, a major insight is that conditional sampling is also affected by privacy mechanisms. This is noticeable in all DPSGD-generated samples, but it occurs at low noise magnitudes for the VAE, as seen in Figure \ref{fig:gen-samples-subfigs-b}. Thus, as the magnitude of the noise increases, the ability of the generator to produce the desired class decreases.

\begin{figure}[!ht]
  \centering

  \begin{subfigure}[b]{0.4\linewidth}
    \centering
    \includegraphics[width=\linewidth]{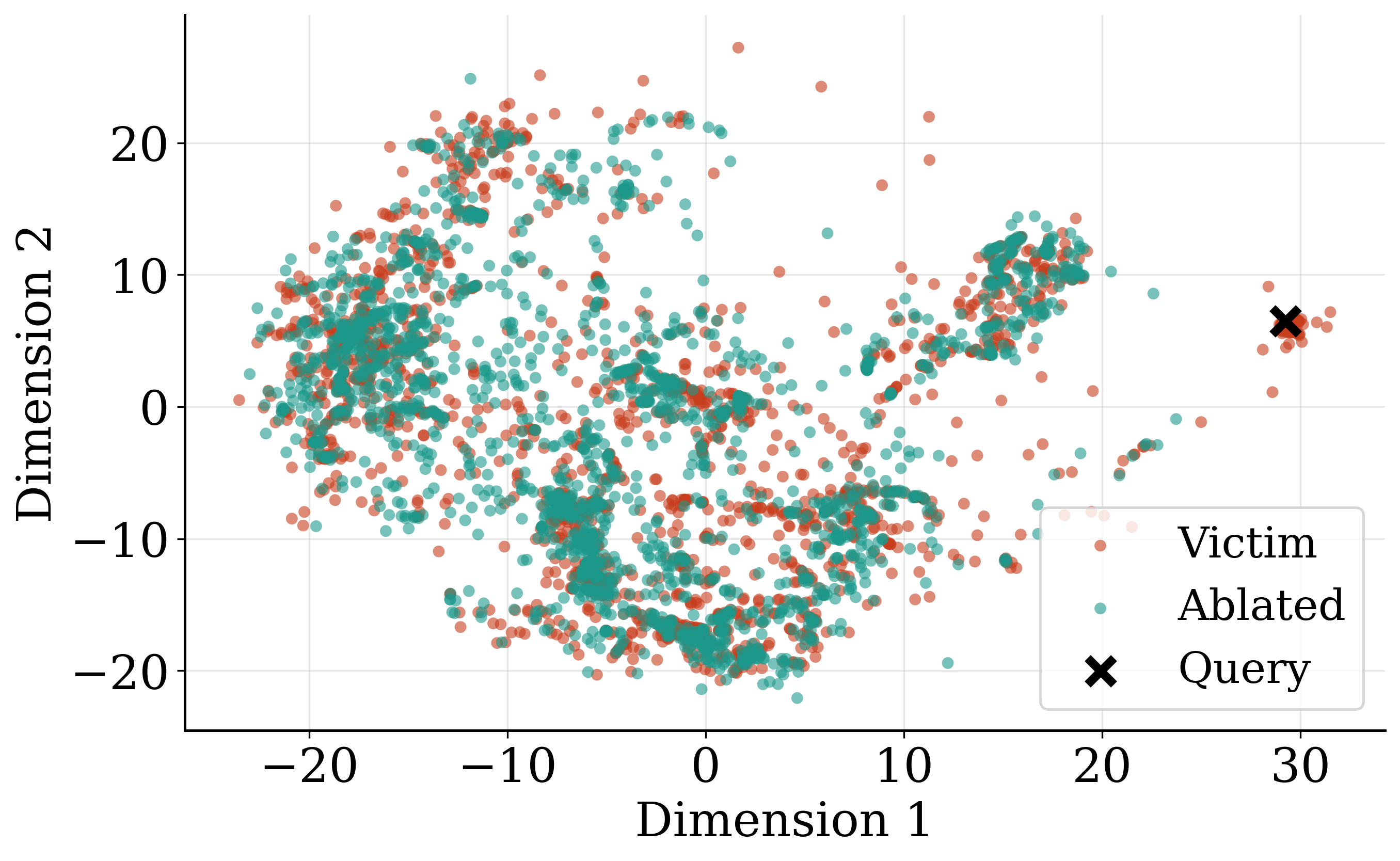}
    \caption{VAE: Isomap projection}
  \end{subfigure}
  \hspace{0.5em}
  \begin{subfigure}[b]{0.4\linewidth}
    \centering
    \includegraphics[width=\linewidth]{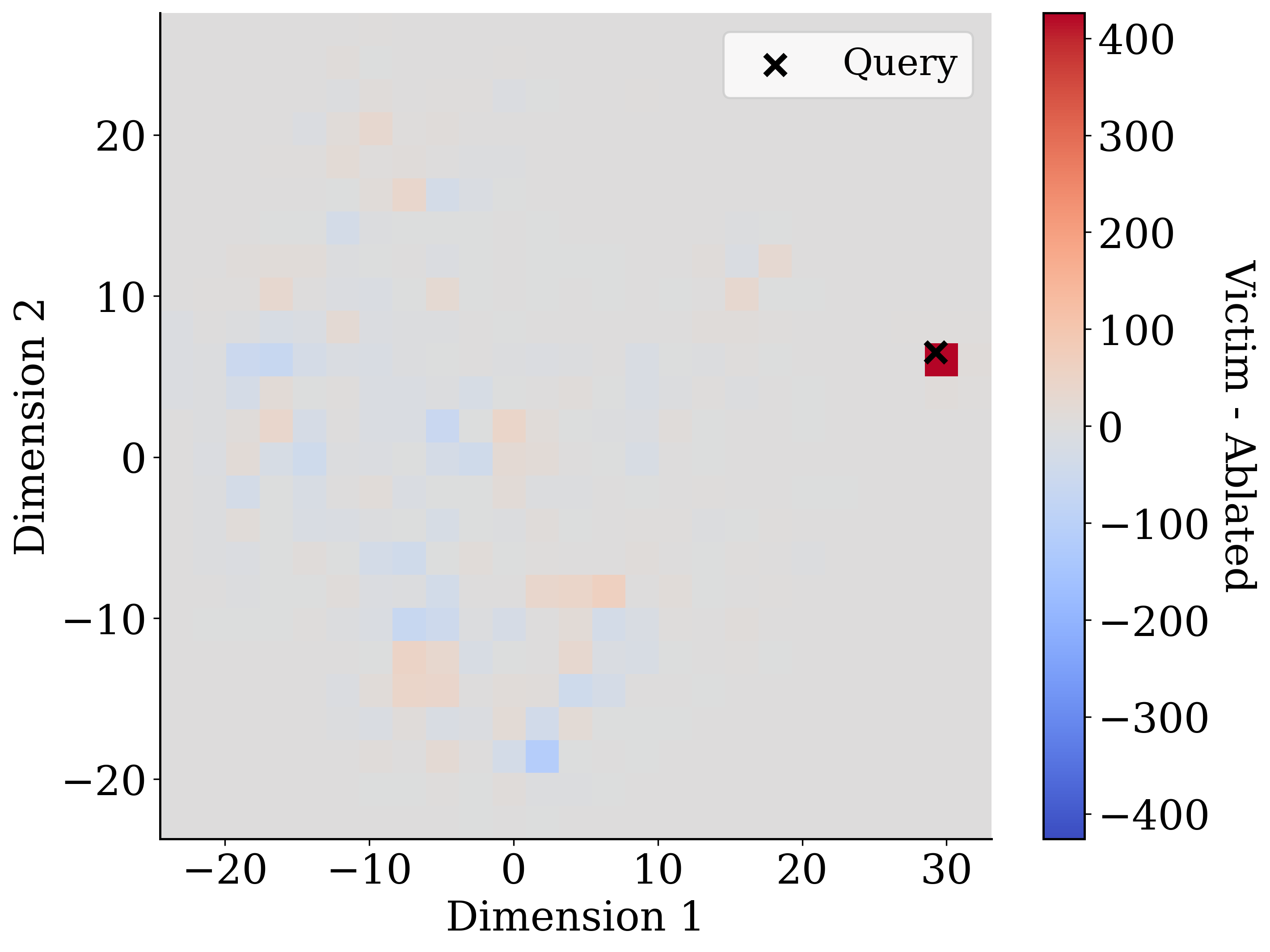}
    \caption{VAE: 2D histogram}
  \end{subfigure}


  \begin{subfigure}[b]{0.4\linewidth}
    \centering
    \includegraphics[width=\linewidth]{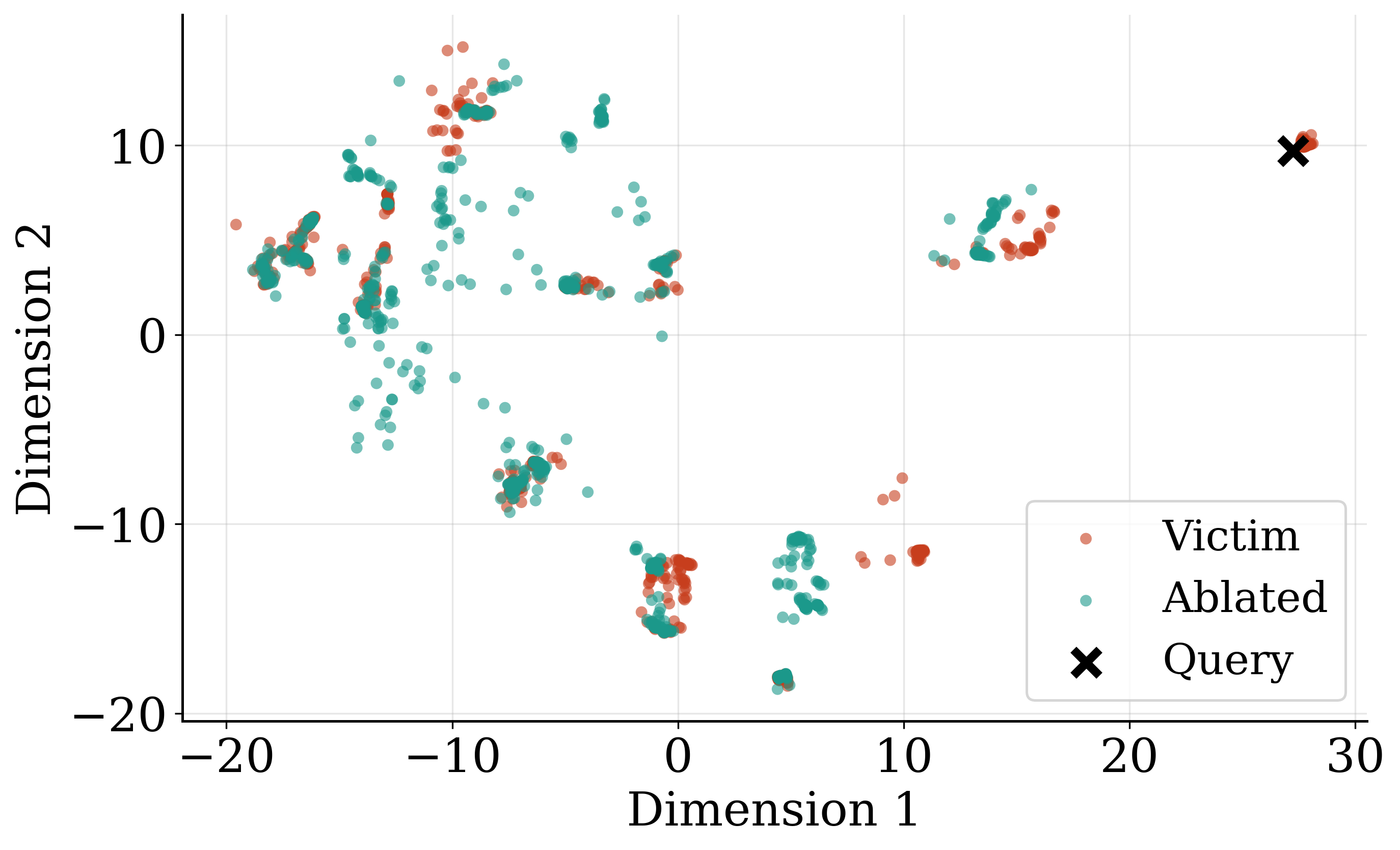}
    \caption{GAN: Isomap projection}
  \end{subfigure}
  \hspace{0.5em}
  \begin{subfigure}[b]{0.4\linewidth}
    \centering
    \includegraphics[width=\linewidth]{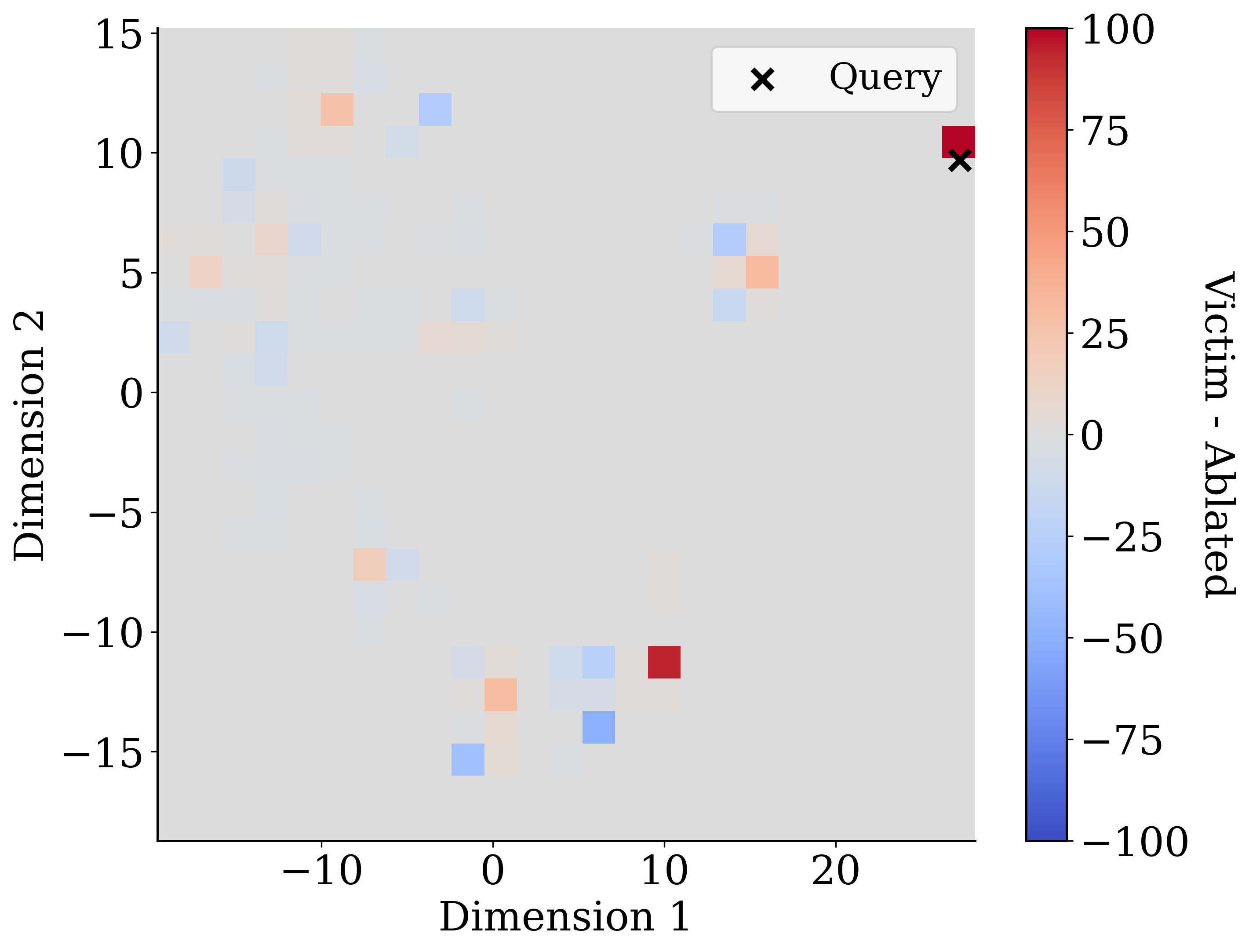}
    \caption{GAN: 2D histogram}
  \end{subfigure}

  \begin{subfigure}[b]{0.4\linewidth}
    \centering
    \includegraphics[width=\linewidth]{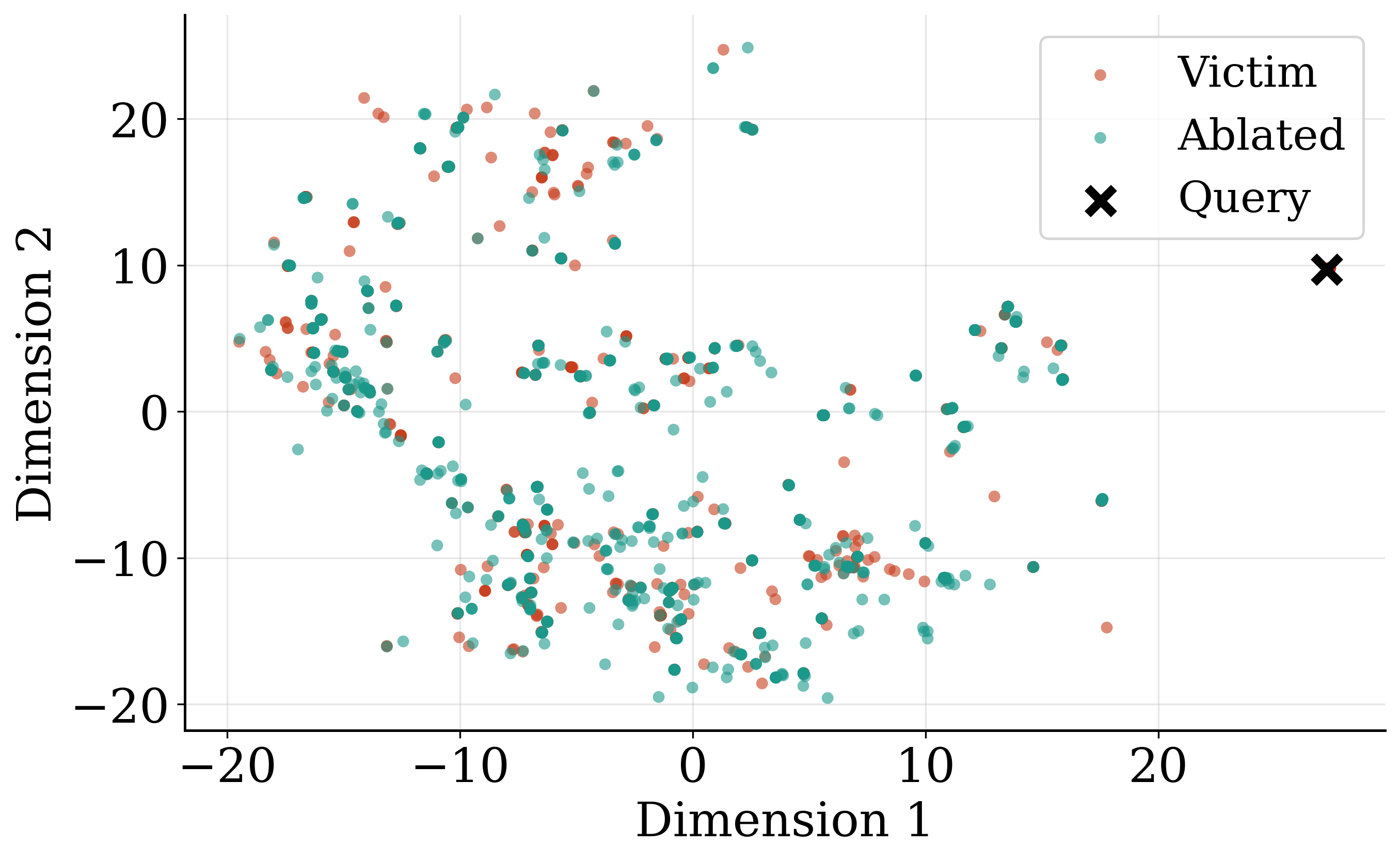}
    \caption{DDPM: Isomap projection}
  \end{subfigure}
  \hspace{0.5em}
  \begin{subfigure}[b]{0.4\linewidth}
    \centering
    \includegraphics[width=\linewidth]{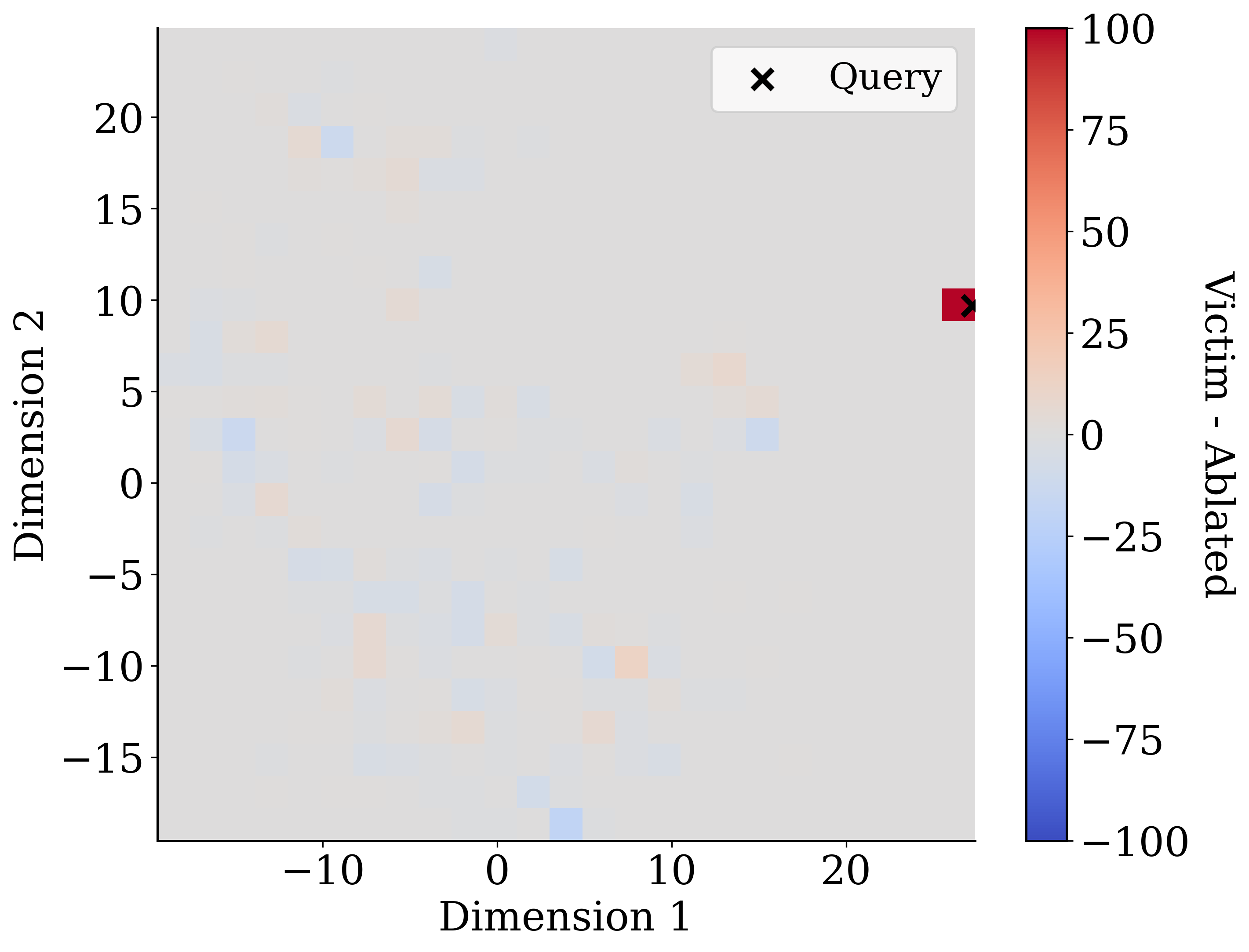}
    \caption{DDPM: 2D histogram}
  \end{subfigure}

  \vspace{0.5em}

  \caption{Projections and 2D histograms for three models at $\sigma = 0.00$. Left column: Isomap embeddings showing the structure of the victim and ablated synthetic sets. Rigth column: local density plots. All models memorise the query sample and generate it during the sampling process, incurring in a high privacy risk, but VAE places more density into the query region resulting in the highest leakage. Moreover, subfigures (c) and (d) show the effects of mode collapse phenomenon associated with GAN. This behaviour reflects the distinct degree of memorisation that each of the architectures suffers under data-scarce conditions.}
  \label{fig:privacy-composite}
\end{figure}

\paragraph{Victim Model Exploration}
To better understand spatial relationships and possible leakage, we visualised latent embeddings from synthetic data using Isomap and 2D histograms. At $\sigma = 0.00$, the three victim models concentrated synthetic samples around the query region. This phenomenon is visible in the right column of Figure \ref{fig:privacy-composite}, where the red pixels near the query indicate a high density of victim-generated samples. This behaviour suggests a serious privacy risk. Moreover, the GAN had serious problems in capturing the full image space and just collapsed into a few modes. Figure \ref{fig:gen-samples-subfigs-c} reflects this behaviour with very densely populated regions located in these modes.

\begin{figure}[!ht]
    \centering
    \includegraphics[width=0.5\linewidth]{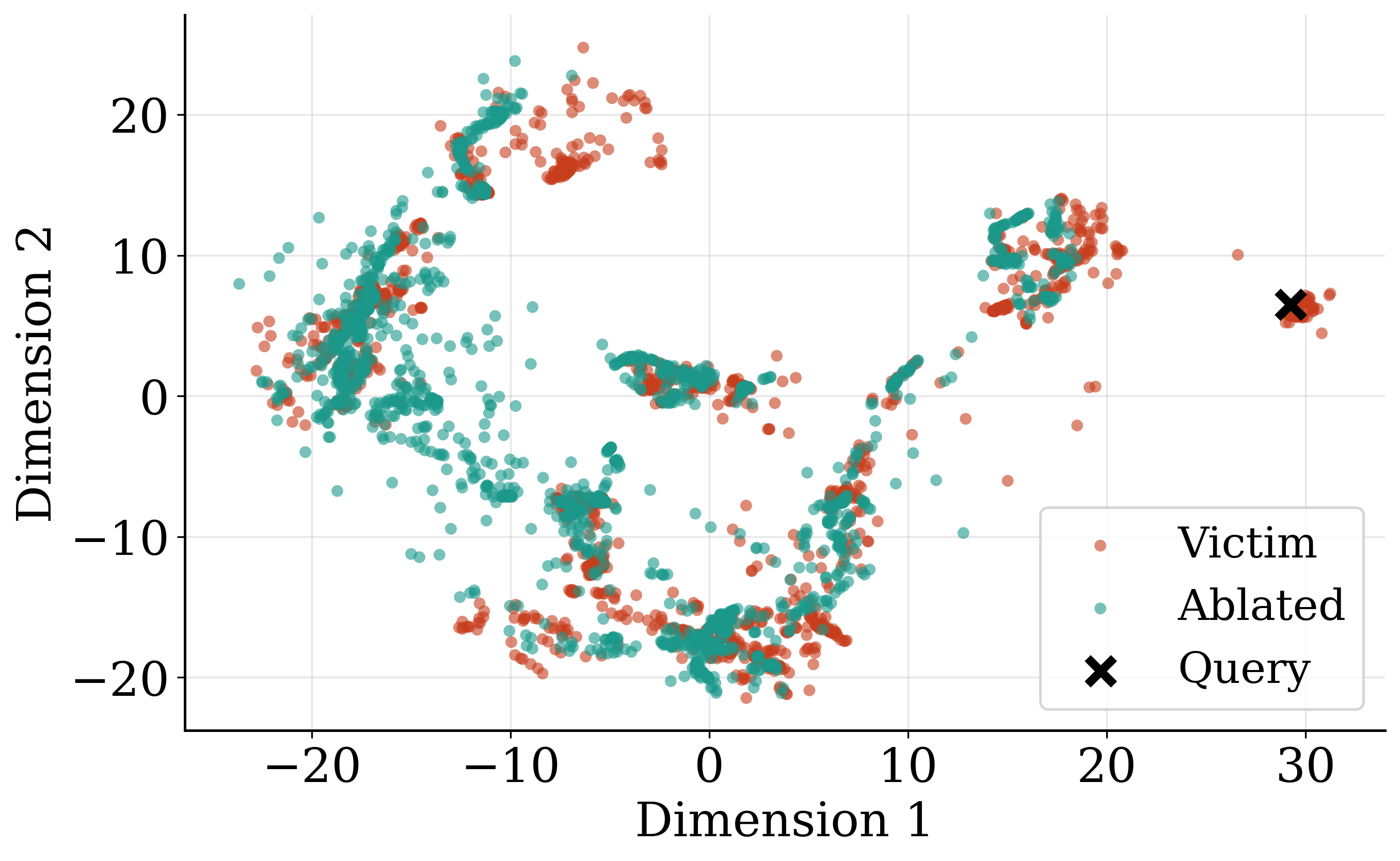}
    \caption{Isomap projection for the GAN model at $\sigma = 0.03$ covering all modes from the original data. In contrast to noiseless training, DPSGD shows a regularising effect which alleviates mode collapse.}
    \label{fig:mnist-gan-scatter-sigma003}
\end{figure}

A key insight from our analysis was that DPSGD improved the GAN’s ability to generalise by promoting broader mode exploration (Figure \ref{fig:mnist-gan-scatter-sigma003}). In the case of SGD, GAN suffered mode collapse, while in the case of DPSGD, mode collapse was mitigated.

This behaviour highlighted the role of DPSGD as an implicit regulariser: by injecting noise into the training process, it prevented the discriminator from overfitting to specific examples, thereby avoiding generator collapse. As a result, the generator was encouraged to explore a broader portion of the data distribution, enhancing coverage and diversity in the synthetic output.

\subsubsection{Quantitative Analysis}

\begin{figure}[!ht]
    \centering
    \begin{subfigure}{0.49\linewidth}
        \centering
        \includegraphics[width=\linewidth]{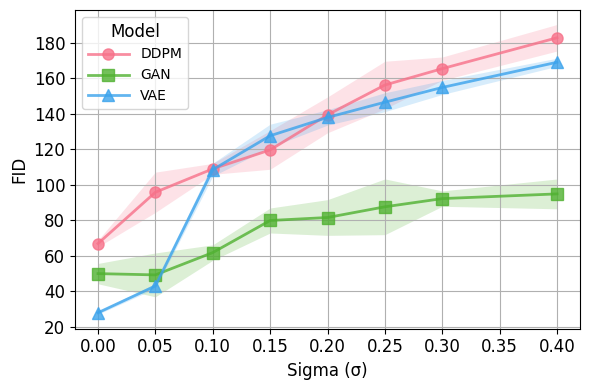}
        \caption{FID. Lower is better.}
        \label{fig:mnist-fid}
    \end{subfigure}
    \begin{subfigure}{0.49\linewidth}
        \centering
        \includegraphics[width=\linewidth]{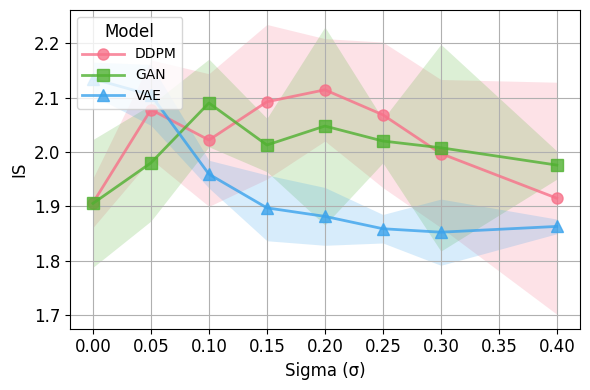}
        \caption{IS. Higher is better.}
        \label{fig:mnist-is}
    \end{subfigure}

    \vspace{2.0em}

    \begin{subfigure}{0.49\linewidth}
        \centering
        \includegraphics[width=\linewidth]{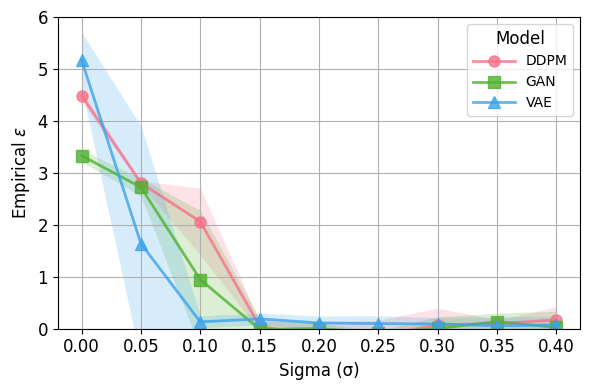}
        \caption{Empirical $\varepsilon$. Lower is better.}
        \label{fig:mnist-empirical-epsilon}
    \end{subfigure}
    \begin{subfigure}{0.49\linewidth}
        \centering
        \includegraphics[width=\linewidth]{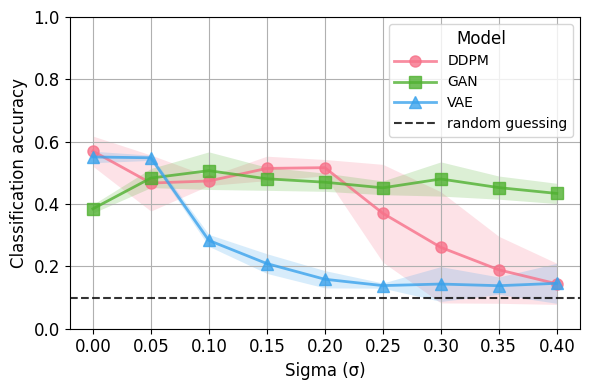}
        \caption{Accuracy. Higher is better.}
        \label{fig:mnist-utility-accuracy}
    \end{subfigure}

    \caption{Comparative results in MNIST measured through FID, IS, empirical $\varepsilon$, and utility accuracy under varying DPSGD noise levels ($\sigma$). Each metric includes 0.9 confidence intervals. GAN seems to be the most robust architecture and noise even seems to alleviate the signs of mode collapse present at $\sigma = 0.00$.}
    \label{fig:mnist-combined}
\end{figure}

Figure \ref{fig:mnist-combined} presents a comparative analysis of generative model performance and privacy at several DPSGD noise levels ($\sigma$) in the MNIST data set. The metrics evaluated include FID and IS for image fidelity and diversity, empirical $\varepsilon$ for privacy leakage, and classification accuracy as a proxy of utility. Among fidelity metrics, FID and IS emerged as the most meaningful and informative indicators because of their well-established correlation with perceptual quality and distributional alignment between real and synthetic samples. Lower FID values indicate closer alignment to the real data distribution, while higher IS reflects both quality and class diversity. These metrics revealed distinct trade-offs among the evaluated models for the MNIST use case. The VAE architecture emphasised privacy at the expense of fidelity and utility, resulting in a rapid decline in these metrics with increased noise levels. In contrast, the DDPM architecture demonstrated a more robust balance for MNIST data, maintaining stable utility until higher sigma values were reached before significant deterioration occurred. Notably, the GAN architecture provided an optimal trade-off in this scenario: even at elevated privacy levels, it consistently retained both high fidelity and strong utility.

\begin{figure}[!ht]
    \centering
    \includegraphics[width=0.6\linewidth]{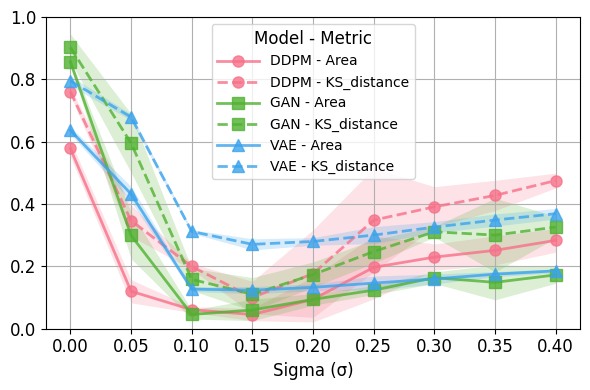}
    \caption{KS distance and normalised area between ECDF curves for several $\sigma$ values with 0.9 confidence intervals. Lower values are better.  $\sigma = 0.15$ provided the best overall results for the three models.}
    \label{fig:mnist-ecdf-ksdist-area}
\end{figure}

\paragraph{Privacy}
The ECDF results presented in Figure \ref{fig:mnist-ecdf-ksdist-area} reveal similar trends for the three architectures.  Depending on the noise, VAE performed worse at medium levels, GAN at low levels, and DDPM at high levels. The best value for the three models is $\sigma = 0.15$, in which DDPM and GAN were more robust than VAE.

Figure \ref{fig:mnist-empirical-epsilon} shows that all three architectures exhibit similar trends in empirical $\varepsilon$ estimation. Low values of $\sigma$ indicate high privacy risks, and all models show rapid improvement until $\sigma \approx 0.15$, where the metric approaches zero. However, the curve slopes reveal that privacy leakage decreases most slowly for DDPM, followed by GAN, and most rapidly for VAE.

Together, these analyses demonstrate that training with DPSGD significantly alters the way generative models learn the original MNIST data distribution. The VAE exhibits more abrupt transitions with noise, whereas the GAN and DDPM gradually adopt safer behaviours. These visual and statistical cues make a compelling case for the latter models to be more resilient to privacy risks in data-limited training scenarios.

\paragraph{Fidelity}
The three models exhibited distinct behaviours under differential privacy constraints. For the VAE, fidelity degraded sharply with the introduction of DPSGD noise. As shown in Table \ref{tab:fid-metrics}, the FID values increased dramatically from 27.97 to 169.04, indicating a substantial loss in perceptual and structural coherence. The IS similarly decreased from 2.13 to 1.86, reflecting reduced image quality and diversity. Interestingly, PSNR values improved slightly with increased noise (from 9.89 to 11.00), suggesting that blurrier outputs paradoxically reduce pixel-level error. Both LPIPS and SSIM remained relatively stable, which may reflect the model's tendency to produce consistent but low-detail images under noisy gradients.

The DDPM model showed intermediate behaviour, with FID degrading from 66.93 to 182.84 and IS remaining relatively stable. However, DDPM exhibited the most concerning trend in structural similarity, with SSIM declining significantly from 0.53 to 0.36, suggesting a substantial loss of image structure under privacy constraints. LPIPS also degraded more than the other models, indicating a reduced perceptual quality.

In contrast, GANs demonstrated superior robustness in all metrics. The FID values increased more moderately from 50.02 to 94.91, while IS remained nearly constant. Both LPIPS and PSNR showed minimal variation, and SSIM maintained stable values around 0.64-0.66, indicating that GAN-generated images retained spatial sharpness and perceptual consistency despite gradient perturbations.

\begin{table}[!ht]
\centering
\begin{tabular}{|c|c|c|c|c|c|}
\hline
\textbf{Metric} & $\sigma=0.0$ & $\sigma=0.1$ & $\sigma=0.2$ & $\sigma=0.3$ & $\sigma=0.4$ \\ \hhline{|======|}
FID (VAE)   & \textbf{27.97}& 108.54& 137.97& 154.86& 169.04\\ \hline
FID (GAN)   & 50.02& \textbf{61.86}& \textbf{81.61}& \textbf{92.26}& \textbf{94.91}\\\hline
 FID (DDPM)& 66.93& 109.02& 139.42& 165.45&182.84\\ \hhline{|======|}
LPIPS (VAE) & \textbf{0.40}& \textbf{0.38}& \textbf{0.38}& \textbf{0.37}   & \textbf{0.37}\\ \hline
LPIPS (GAN) & \textbf{0.40}& 0.41   & 0.41   & 0.41   & 0.41\\\hline
 LPIPS (DDPM)& 0.42& 0.45& 0.46& 0.46&0.47\\ \hhline{|======|}
PSNR (VAE)  & \textbf{9.89}& \textbf{10.47}& \textbf{10.72}& \textbf{10.90}& \textbf{11.00}\\ \hline
PSNR (GAN)  & 9.79& 9.72& 9.75& 9.64& 9.64\\ \hline
 PSNR (DDPM)& 9.48& 9.21& 9.31& 9.49&9.45\\ \hhline{|======|}
 IS (VAE)& \textbf{2.13}& 1.96& 1.88& 1.85&1.86\\ \hline
 IS (GAN)& 1.91& \textbf{2.09}& 2.05& \textbf{2.01}&\textbf{1.98}\\ \hline
 IS (DDPM)& 1.91& 2.02& \textbf{2.11}& 2.00&1.91\\ \hhline{|======|}
 SSIM (VAE)& 0.65& \textbf{0.65}& \textbf{0.65}& 0.64&0.62\\ \hline
 SSIM (GAN)& \textbf{0.66}& 0.64& 0.64& \textbf{0.65}&\textbf{0.65}\\ \hline
 SSIM (DDPM)& 0.53& 0.42& 0.40& 0.38&0.36\\ \hline
\end{tabular}
\caption{Fidelity metrics (FID, lower better; LPIPS, lower better; PSNR, higher better; IS, higher better; and SSIM, higher better) for synthetic images generated by the three architectures under varying DPSGD noise levels. Best-performing scores per metric and noise setting are highlighted in bold. Depending on the metric, a different model is the best performing, so a fine-grained analysis is required to address the most suitable metrics, depending on the problem and data at hand. In fact, LPIPS and PSNR may behave counterintuitively due to the overly smoothed outputs typically produced by VAEs \cite{guo_grey-box_2025}.}
\label{tab:fid-metrics}
\end{table}

The two most promising metrics for this case, FID and IS, are shown in Figures \ref{fig:mnist-fid} and \ref{fig:mnist-is} with 0.9 confidence intervals. All of these results reveal fundamental distinctions among the three generative architectures under differential privacy constraints for the MNIST case. While VAE achieves the best initial fidelity, it rapidly degrades under DPSGD noise. GAN maintains more stable performance across all metrics, making it particularly suitable when consistent synthetic data quality is essential. DDPM shows intermediate degradation patterns but exhibits loss of structural similarity, suggesting potential limitations for applications requiring preserved image structure. Consequently, GAN may offer the most robust architecture under DP constraints when synthetic data fidelity is fundamental.

\paragraph{Utility}
Utility was evaluated by training classifiers on synthetic data generated by the three models and measuring their accuracy on a holdout test set. The classification task was chosen as the downstream utility benchmark due to MNIST's natural labelling structure. The results are shown in Figure \ref{fig:mnist-utility-accuracy}.

The VAE struggled significantly as the DPSGD noise increased. Accuracy dropped from approximately 0.6 (at $\sigma=0.0$) to near random guessing levels around 0.15 for higher noise values. This decline was expected: the scarcity of real training samples and the vulnerability of VAE to DPSGD meant that the synthetic data lacked the diversity and quality needed for generalisation.

In contrast, the GAN showed a surprising trend: utility improved under moderate noise. Accuracy increased from 0.40 (at $\sigma=0.0$) to a plateau of 0.50 at $\sigma \in [0.10, 0.30]$. This effect likely stems from DPSGD acting as a regulariser on the generator, smoothing over memorised patterns, and promoting greater variability in output. As a result, classifiers trained on these GAN-generated data were more effective.

In the middle of these situations, the DDPM exhibited a stable behaviour at $\sigma \leq 0.20$, while for larger noise values, the performance of the downstream classifier rapidly degraded.

\begin{figure}[!ht]
    \centering
    \includegraphics[width=0.6\linewidth]{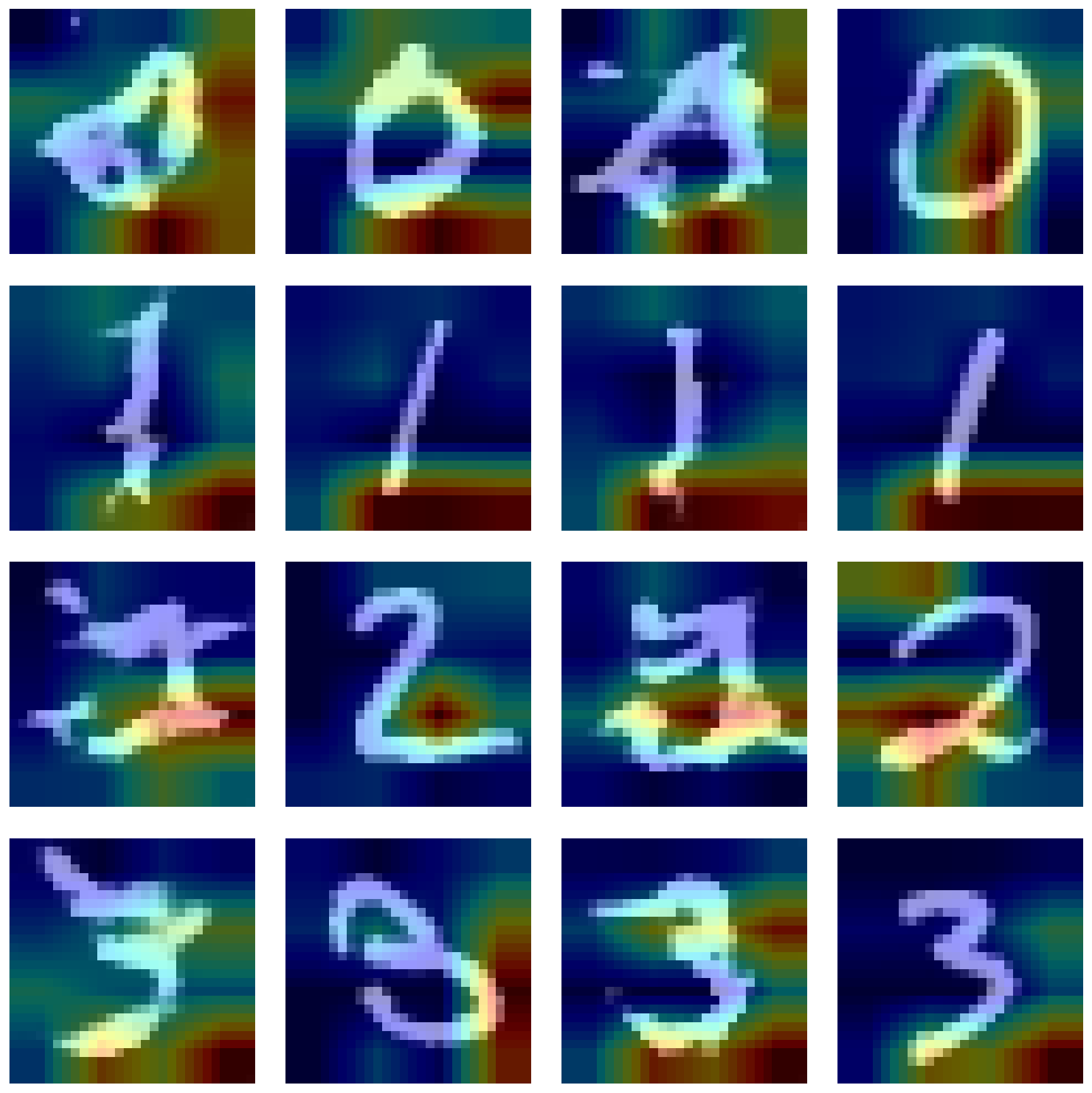}
    \caption{GradCAM visualisations over real data from a CNN classifier trained on GAN-generated data at $\sigma=0.20$. Each pair of images is composed of a synthetic sample (left) and a real sample (right) analysed through the same classifier trained solely on the synthetic samples. The classifier focuses on discriminative spatial regions such as angles, end strokes, and digit curvature.}
    \label{fig:mnist-gradcam-utility}
\end{figure}

To further interpret classifier behaviour, we analysed GradCAM \cite{selvaraju_grad-cam_2017} visualisations of the classifier trained on synthetic data produced by a GAN at $\sigma=0.20$. Figure \ref{fig:mnist-gradcam-utility} shows that the classifier focusses on spatially meaningful regions of the digits. We expected that the focus region would be similar for both synthetic and real data. For instance, in class 0, it attends to a broad region generally located outside the stroke, while for class 1, it concentrates on the lower digit closure. These focused activation zones demonstrate that the classifier has learnt discriminative spatial patterns from the synthetic data that are similarly applied to real data. In addition, it is important to note that, except for class 2, the GAN was able to produce meaningful images, although human perception may not be aligned with automatic pattern recognition, as multiple strokes may make these patterns counterintuitive to us.

\subsection{Application to Medical Data} \label{sec:medical-data}
The previous section aimed to show how the quality of synthetic data, measured through the trade-off between fidelity, privacy, and utility, was affected by the scarcity of training data and by privacy protection mechanisms. However, the MNIST dataset may not fall within the scope of information confidentiality. In contrast, medical data are sensitive by definition, so this section shows a real use case in which privacy must be considered by default.

Medical data fall into the category of sensitive information, and special consideration should be given to privacy guarantees when performing modelling tasks on it, particularly if learning complex data distributions. Deep generative models excel at learning complex distributions but are severely affected by overfitting due to the scarcity of training data. Therefore, all risk conditions are met: scarcity, sensitivity, and complexity, the latter affecting both the model through the degrees of freedom and the data due to its high dimensionality.

\subsubsection{Introduction to the data sets}
Data considered for medical analysis were a subset of the MedMNIST catalogue \cite{yang_medmnist_2023}. From this catalogue, two data sets were selected: OCTMNIST and OrganAMNIST. The former is characterised by retinal OCT samples labelled with 4 classes, while the latter includes abdominal CT samples with 11 classes. Both data sets have the shape $(1,28,28)$, suggesting one channel and 28x28 images. For more information, see the online \href{https://medmnist.com/}{documentation}.

\begin{figure}[!ht]
  \centering
  \includegraphics[width=0.75\linewidth]{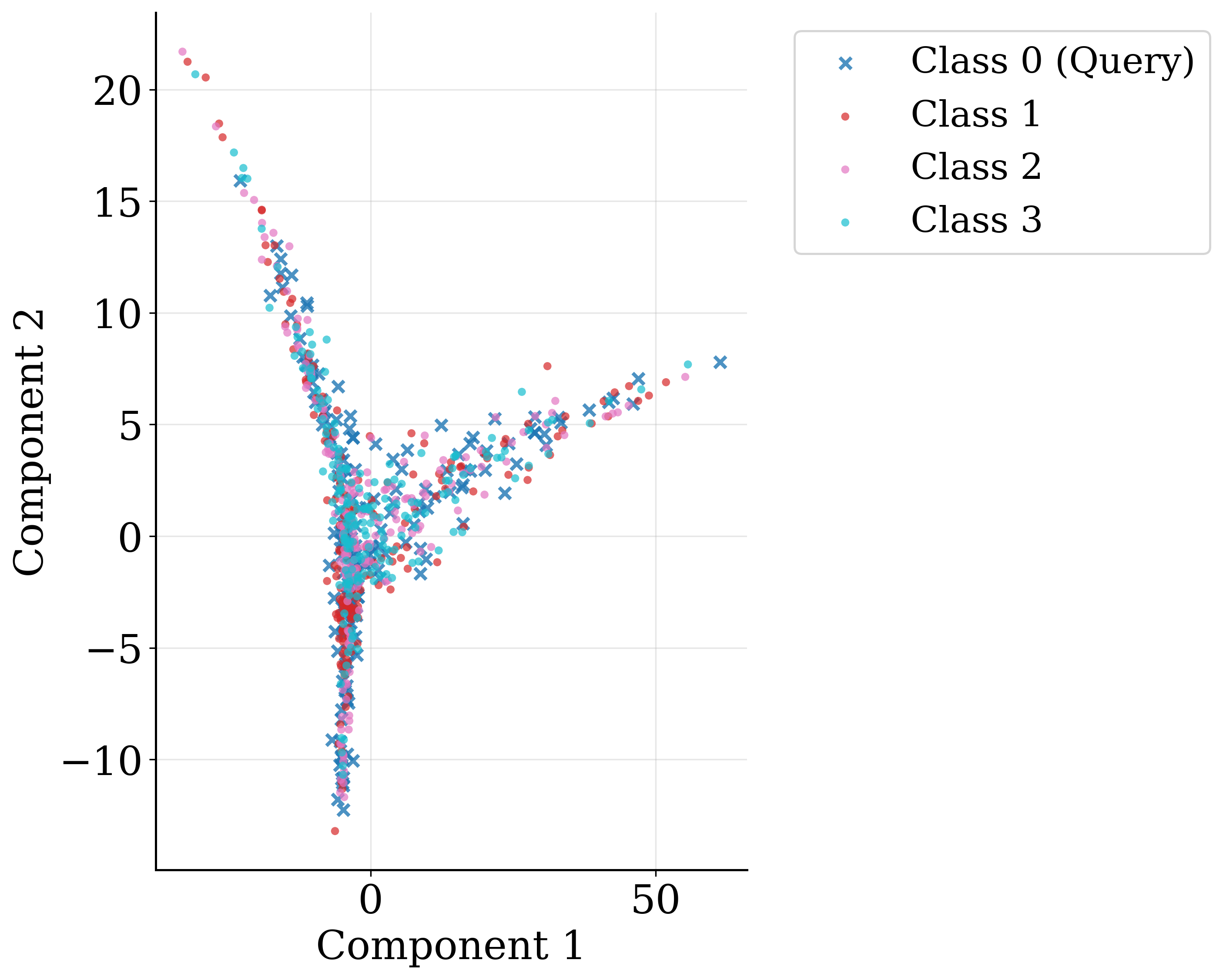}
  \caption{Isomap projections for the OCTMNIST data set with the corresponding labels. Class 0 was selected as the query class.}
  \label{fig:octmnist-projections}
\end{figure}

\begin{figure}[!ht]
  \centering
  \includegraphics[width=0.75\linewidth]{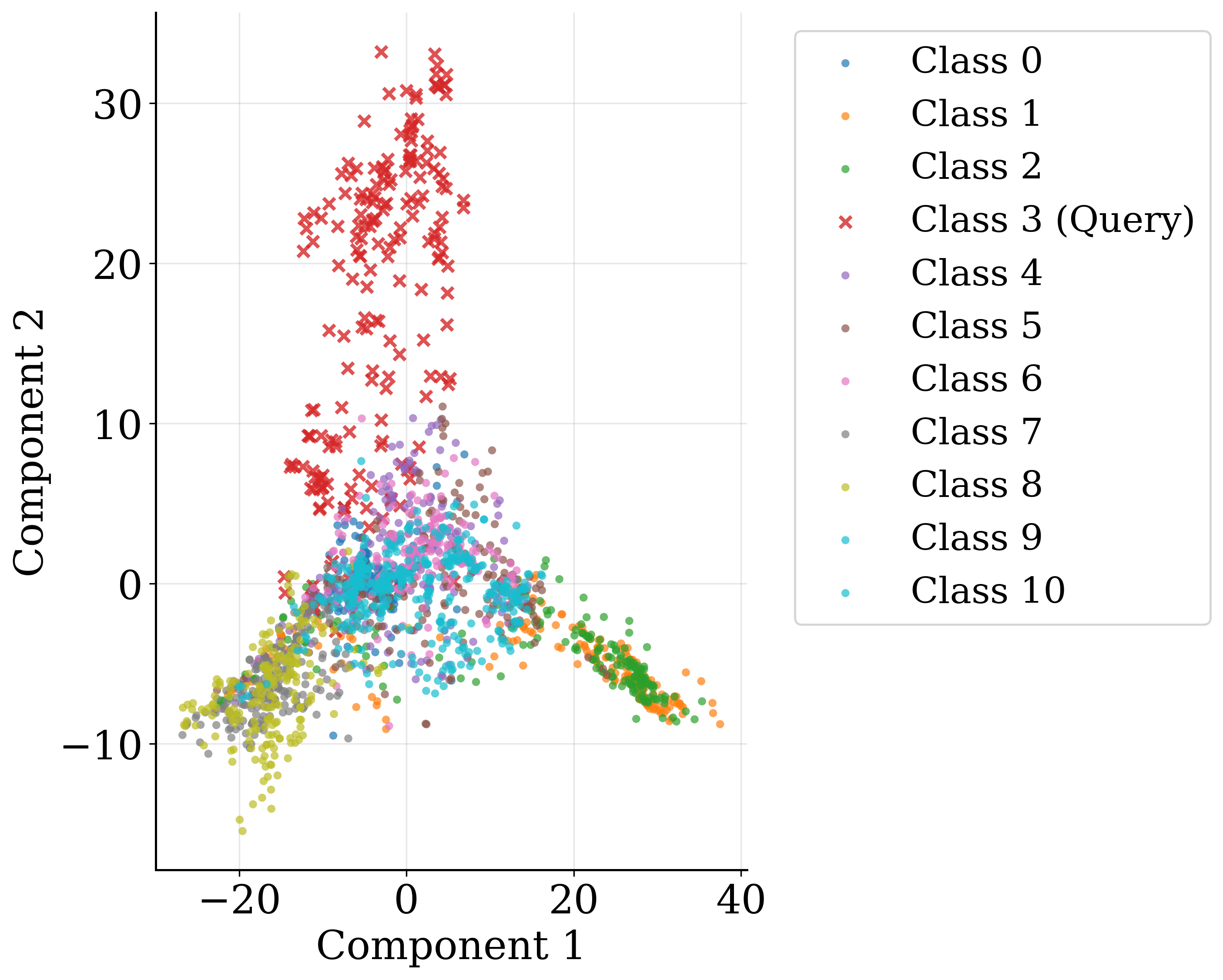}
  \caption{Isomap projections for the OrganAMNIST data set with the corresponding labels. Class 3 was selected as the query class.}
  \label{fig:organamnist-projections}
\end{figure}

We applied the Isomap dimensionality reduction technique to project the data and obtained the structures shown in Figures \ref{fig:octmnist-projections} and \ref{fig:organamnist-projections}. The resulting layout revealed that some classes were clearly distinguishable and naturally formed visual clusters, indicating significant interclass dissimilarity. This structure was exploited to select a strategically appropriate missing class or a worst-case scenario for privacy, namely, when a query data point belongs to a space region with few or no representative samples. In real settings, data scarcity and heterogeneity could produce similar conditions in which information leakage is a concern.

Based on this rationale, we selected classes 0 and 3 to represent the query class for the OCTMNIST and OrganAMNIST datasets, respectively. The following sections describe the results obtained for each of these data sets.

\subsubsection{OCTMNIST Analysis}
To evaluate fidelity and utility, the model was presented with 10 samples per class, giving a total amount of 40 samples. Moreover, for privacy estimation, the selected query class was the 0 label. Hence, the victim model was fed 10 samples per class, except for class 0, which only provided one sample. In contrast, the ablated model was trained without the 0 class sample. Three seeds per setting are run, that is, the combination of model architectures and noise magnitudes, to achieve representative results.

\begin{figure}[!ht]
    \centering
    \includegraphics[width=0.9\linewidth]{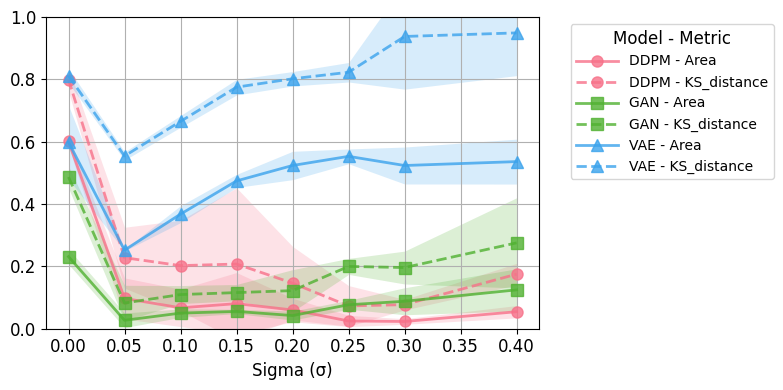}
    \caption{KS distance and normalised area between ECDF curves for several $\sigma$ values for OCTMNIST data set. Lower values are better. All models suffered from a high drift between the ECDF curves at $\sigma = 0.00$ but VAE extended this behaviour along all privacy settings. Moreover, DDPM model suffered from very high variance, which is a sign of adequate and robust learning.}
    \label{fig:octmnist-ecdf-ksdist-area}
\end{figure}

\begin{figure}[!ht]
    \centering
    \begin{subfigure}{0.49\linewidth}
        \centering
        \includegraphics[width=\linewidth]{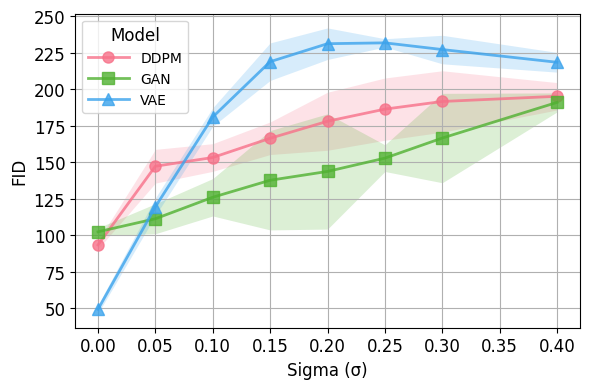}
        \caption{FID. Lower is better.}
        \label{fig:octmnist-fid}
    \end{subfigure}
    \begin{subfigure}{0.49\linewidth}
        \centering
        \includegraphics[width=\linewidth]{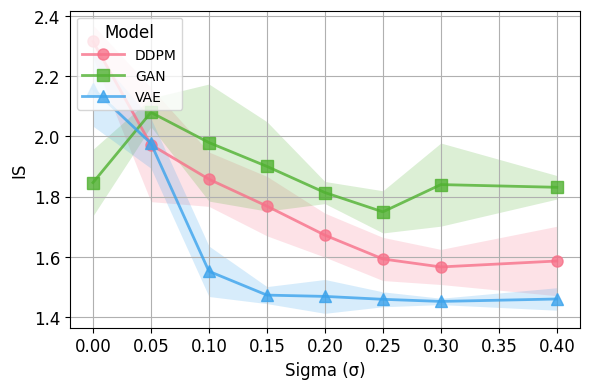}
        \caption{IS. Higher is better.}
        \label{fig:octmnist-is}
    \end{subfigure}

    \begin{subfigure}{0.49\linewidth}
        \centering
        \includegraphics[width=\linewidth]{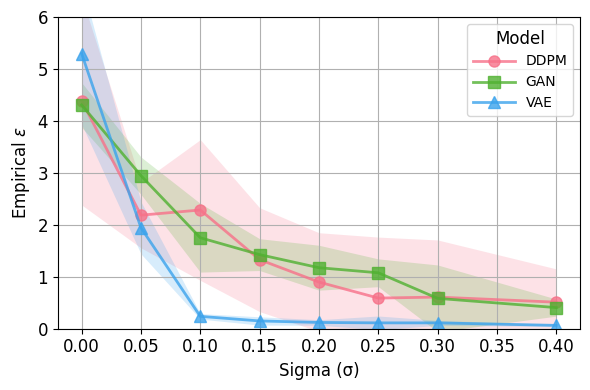}
        \caption{Empirical $\varepsilon$. Lower is better.}
        \label{fig:octmnist-empirical-epsilon}
    \end{subfigure}
    \begin{subfigure}{0.49\linewidth}
        \centering
        \includegraphics[width=\linewidth]{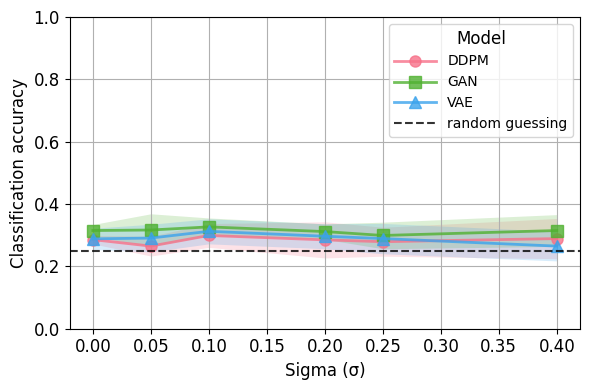}
        \caption{Accuracy. Higher is better.}
        \label{fig:octmnist-utility-accuracy}
    \end{subfigure}

    \caption{Comparative results in OCTMNIST across FID, IS, empirical $\varepsilon$, and utility accuracy under varying DPSGD noise levels ($\sigma$). Each metric includes 0.9 confidence intervals. GAN seems to be the most robust architecture, although none of the models could produce usable samples for classification.}
    \label{fig:octmnist-combined}
\end{figure}

The analysis of the ECDF curves, shown in Figure \ref{fig:octmnist-ecdf-ksdist-area}, demonstrated how the GAN and DDPM models outperformed VAE in all settings. These results aligned with the behaviour observed in the MNIST use case. However, the VAE architecture, because of the enormous KS distance and area between the ECDF curves, either suffered from overfitting or was unable to fit the data. The complementary results in Figure \ref{fig:octmnist-combined} proved the latter.

The OCTMNIST evaluation highlighted distinct behaviours between generative models with varying levels of DP, as illustrated in Figures \ref{fig:octmnist-combined}. In terms of fidelity, Figure \ref{fig:octmnist-fid} showed that the VAE achieved the lowest FID score ($\approx 50$) when trained with standard SGD, indicating high-quality reconstructions. However, its performance degraded sharply once DPSGD was introduced, becoming the model that performed the worst as the noise increased. In contrast, both GAN and DDPM started with higher FID values around 100 under non-private training, but exhibited a more gradual deterioration with increasing $\sigma$. In particular, the GAN demonstrated greater robustness, maintaining relatively lower FID scores across privacy settings. Regarding diversity, Figure \ref{fig:octmnist-is} revealed that both VAE and DDPM initially achieved higher IS ($\approx$ 2.2) under vanilla SGD. However, this advantage dissipated rapidly with the introduction of noise, as the GAN surpassed both models beyond $\sigma \geq 0.05$, achieving stable IS values between 1.8 and 2.0. This suggested that GANs were more capable of preserving sample diversity under privacy constraints.

The privacy leakage, as shown in Figure \ref{fig:octmnist-empirical-epsilon}, further differentiated the models. VAE was quickly overwhelmed by even modest noise levels and was unable to retain query-relevant information beyond $\sigma = 0.10$. In contrast, DDPM and GAN retained more information about the training samples even at higher $\sigma$ values, although this may have implied a less strict adherence to privacy preservation. Finally, as depicted in Figure \ref{fig:octmnist-utility-accuracy}, all models performed poorly in terms of downstream classification utility. The synthetic data generated in all privacy settings led to classifier accuracies near the level of random guessing, indicating that none of the models was able to produce synthetic samples relevant to the task with meaningful predictive power in this setting.

\subsubsection{OrganAMNIST Analysis}
To evaluate fidelity and utility, the model was presented with 10 samples per class, giving a total of 110 samples. Moreover, for privacy estimation, the selected query class was the 0 label. Hence, the victim model was fed 10 samples per class, except for class 0, which only provided one sample. In contrast, the ablated model was trained without the 0 class sample. Three seeds per setting are run, that is, the combination of model architectures and noise magnitudes, to achieve representative results.

\begin{figure}[!ht]
    \centering
    \includegraphics[width=0.75\linewidth]{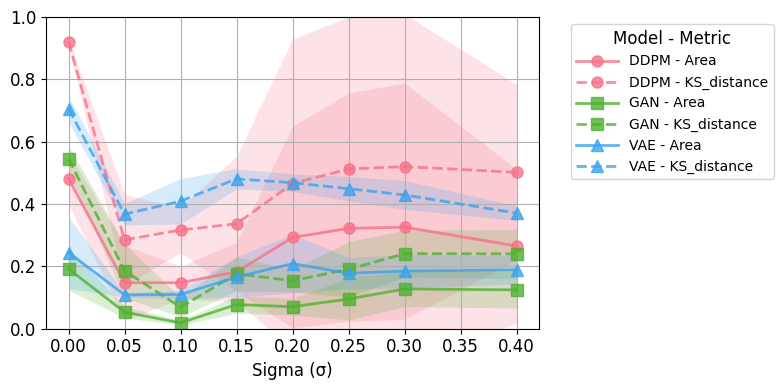}
    \caption{KS distance and normalised area between ECDF curves for several $\sigma$ values for the OrganAMNIST data set. Lower values are better. $\sigma = 0.10$ provided the best results for all models. All models suffered from a high drift at $\sigma = 0.00$. Depending on the amount of noise, either VAE or DDPM was the worse architecture. GAN provided the best results for privacy-constrained settings.}
    \label{fig:organamnist-ecdf-ksdist-area}
\end{figure}

The analysis of the ECDF curves, presented in Figure \ref{fig:organamnist-ecdf-ksdist-area}, showed a trend consistent with previous datasets. Although OrganAMNIST was structurally different, containing more dense and organic visual patterns, all models exhibited similar distributional drift characteristics under varying DPSGD noise levels. Specifically, when no noise was applied ($\sigma = 0.00$), the synthetic data diverged significantly from the real distribution, indicating overfitting. As the noise increased, the alignment between the synthetic and real ECDFs improved. Across all models, $\sigma = 0.10$ yielded the lowest distances, suggesting a good spot between privacy and generalisation. Although the worst-performing model varied depending on the noise level, the GAN consistently provided the most favourable results.

\begin{figure}[!ht]
    \centering
    \begin{subfigure}{0.49\linewidth}
        \centering
        \includegraphics[width=\linewidth]{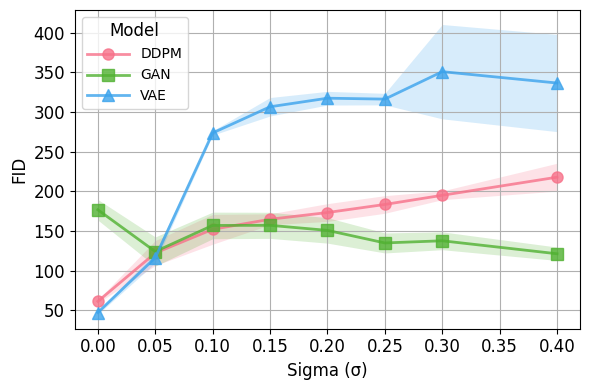}
        \caption{FID. Lower is better.}
        \label{fig:organamnist-fid}
    \end{subfigure}
    \begin{subfigure}{0.49\linewidth}
        \centering
        \includegraphics[width=\linewidth]{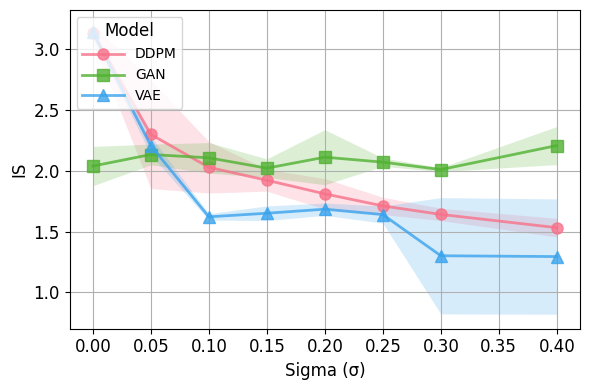}
        \caption{IS. Higher is better.}
        \label{fig:organamnist-is}
    \end{subfigure}

    \begin{subfigure}{0.49\linewidth}
        \centering
        \includegraphics[width=\linewidth]{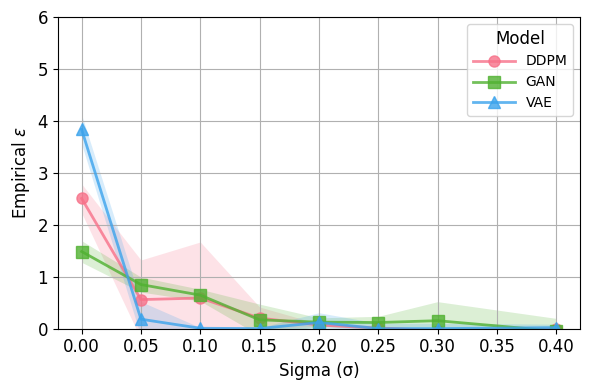}
        \caption{Empirical $\varepsilon$. Lower is better.}
        \label{fig:organamnist-empirical-epsilon}
    \end{subfigure}
    \begin{subfigure}{0.49\linewidth}
        \centering
        \includegraphics[width=\linewidth]{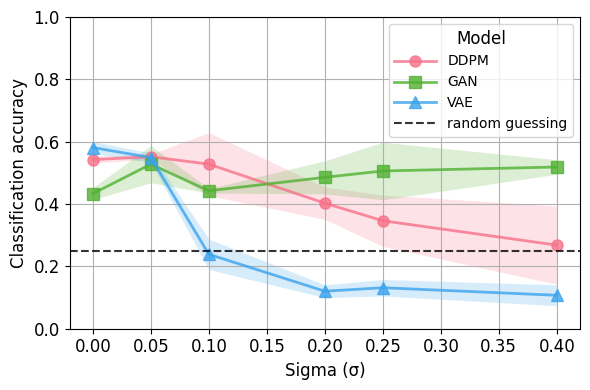}
        \caption{Accuracy. Higher is better.}
        \label{fig:organamnist-utility-accuracy}
    \end{subfigure}

    \caption{Comparative results in OrganAMNIST across FID, IS, empirical $\varepsilon$, and utility accuracy under varying DPSGD noise levels ($\sigma$). Each metric includes 0.9 confidence intervals. GAN seems to be the most robust architecture.}
    \label{fig:organamnist-combined}
\end{figure}

Inspecting Figure \ref{fig:organamnist-combined}, the models offered different behaviours across privacy settings. Given a non-private training mode, DDPM and VAE were very performant on both fidelity (FID $\approx 50$, IS $\approx 3$) and utility, measured through accuracy, with a value of approximately 0.6. In contrast, GAN offered worse performance with FID $\approx 175$, IS $\approx 2.0$. However, in private settings, VAE rapidly degraded and provided the worst results across all $\sigma$ magnitudes for both utility and privacy. In contrast, GAN was robust for all the $\sigma$ values analysed. DDPM offered middle results: lower $\sigma$ were easily handled by the model, but $\sigma \geq 0.10$ had a huge impact on model performance. In terms of privacy, VAE rapidly became private, even at $\sigma \approx 0.05$, while DDPM and GAN became private at $\sigma \approx 0.15$.

\section{Discussion}
The case studies showed clear differences among the three generative models in how they responded to limited data and privacy constraints. Across all datasets, VAE quickly lost image quality and usefulness when noise increased due to DPSGD. In contrast, GAN was more robust and maintained higher quality and usefulness, particularly in MNIST and OrganAMNIST. DDPM performed between these two models, showing better stability and quality than VAE, but still behind GAN. Thus, GAN appears to be best suited for balancing privacy and quality, although careful monitoring of its actual privacy leakage remains necessary.

In OCTMNIST, all models had poor results when training classifiers on synthetic data, mainly because of the data scarcity and complexity. The classifier's accuracy was close to random guessing. This highlights a key limitation: synthetic data from small and complex data sets can only offer limited practical utility. Therefore, realistic privacy and utility goals must consider the size of the dataset and the complexity of the task.

An important finding is that the utility of synthetic data strongly depends on whether the conditional distributions of the original data are preserved. Even if synthetic data perform well in one downstream task (like classification), it may not perform similarly in another task because different tasks impose distinct requirements.

Overall, these results underline the fundamental trade-off between privacy, quality, and utility in synthetic data generation. Improving one dimension usually compromises others, consistent with the no free lunch principle. Therefore, selecting the right generative model should depend on the specific privacy and usability needs of each application. Additionally, data scarcity increases privacy risks because generative models can memorise data points, exposing sensitive information. Both $\sigma$ and the generative modelling options should be analysed in the three dimensions to find an optimal trade-off. However, we recommend applying the principle of least privilege by fixing the desired amount of utility or fidelity.

Interestingly, using DPSGD sometimes helped GANs produce more varied and useful samples, particularly in the MNIST and OrganAMNIST cases. This suggests that privacy methods might indirectly improve GAN generalisation by acting as a regularizer and reducing mode collapse, thus highlighting that, under certain circumstances, privacy constraints can lead to improved outcomes without necessarily breaking the no free lunch principle.

Note that DPSGD was applied differently to GAN compared to DDPM and VAE. While the latter two were affected throughout their architectures, only the discriminator of GAN was trained with DPSGD, following established practices. This subtle difference may explain the superior performance of GAN and must be considered when interpreting these results.

Finally, evaluating privacy through likelihood-based methods is challenging in high-dimensional and small-data settings. Techniques like Isomap can offer useful alternatives by reducing data dimensions and preserving key relationships, but their effectiveness depends on the specific problem and data structure.

\section{Conclusions and Future Directions}
This study introduced an empirical framework specifically designed to evaluate synthetic image generation under data scarcity and differential privacy constraints. Applying this framework revealed that GAN frequently maintained higher fidelity and utility compared to DDPM and VAE across various scenarios. DDPM performed moderately, while VAE rapidly degraded under privacy constraints. The framework also demonstrated significant limitations of synthetic data utility in small and complex datasets such as OCTMNIST, highlighting the importance of conditional distribution preservation for downstream tasks. Furthermore, the findings suggested that DPSGD can sometimes improve GAN generalisation, emphasising the need for careful privacy monitoring. These insights validate the proposed evaluation approach, particularly for regulated domains such as medical imaging, and highlight the importance of balancing fidelity, privacy, and utility in synthetic data generation.

Future research should focus on bridging regulatory and technical perspectives, clearly defining how much privacy is needed and achievable. Practical frameworks are needed to effectively balance privacy with data usefulness. Extending the analyses to other types of data and generative models will help to better understand the strengths and limitations of synthetic data generation in each scenario. Additionally, developing evaluation methods tailored to data scarcity and multiple quality dimensions (privacy, fidelity, utility) is crucial. Because privacy risks increase with more complex data distributions, a comprehensive review of current techniques may be necessary to advance the domain of synthetic data generation.

More specifically, future research directions should prioritise the development of models tailored explicitly to particular tasks by adhering to the principle of least privilege. Such models would aim to maximise the mutual information of relevant features and data structures while effectively masking non-essential information. This approach could potentially yield better trade-offs by incurring in lower privacy risks, thus aligning more robustly with privacy compliance requirements.

In addition, condensing the three dimensions into a single metric seems reasonable to provide a harmonised criterion. To achieve it, several limitations need to be addressed, including the difference in metric domains, scale, and ordering.

\section*{Declaration}
\subsection*{Availability of data and materials}
All data referenced in this article is open source. It includes MNIST and MedMNIST data sets. The code is available in the Github repository \url{https://github.com/BorjaArroyo/synthetic-images-tradeoff}

\subsection*{Competing interests}
The authors declare that they have no known competing financial interests or personal relations that could have appeared to influence the work reported in this article.

\subsection*{Funding}
This work was supported by Synthetic Generation of Hematological Data over Federated Computing Frameworks (SYNTHEMA) project from Horizon Europe under Grant 101095530. Views and opinions expressed are, however, those of the authors only and do not necessarily reflect those of the European Union or the European Commission. Neither the European Union nor the granting authority can be held responsible for them.

\subsection*{Authors' contributions}
\textbf{Borja Arroyo Galende}: Conceptualization, Methodology, Software, Writing – original draft. \textbf{Alejandro Almodóvar}: Methodology, Validation, Writing – review \& editing. \textbf{Patricia A. Apellániz}: Methodology, Validation, Writing – review \& editing. \textbf{Juan Parras}: Resources, Supervision, Writing – review \& editing, Conceptualization. \textbf{Silvia Uribe}: Resources, Funding acquisition, Supervision, Project administration. \textbf{Santiago Zazo}: Resources, Funding acquisition, Supervision, Project administration, Conceptualization.

\section*{Acknowledgements}
This work was supported by Synthetic Generation of Hematological Data over Federated Computing Frameworks (SYNTHEMA) project from Horizon Europe under Grant 101095530. Views and opinions expressed are, however, those of the authors only and do not necessarily reflect those of the European Union or the European Commission. Neither the European Union nor the granting authority can be held responsible for them.

\section*{Declaration of competing interests}
The authors declare that they have no known competing financial interests or personal relations that could have appeared to influence the work reported in this article.

\section*{Availability of data and materials}
All data referenced in this article is open source. It includes MNIST and MedMNIST data sets. The code is available in the Github repository \url{https://github.com/BorjaArroyo/synthetic-images-tradeoff}

\bibliographystyle{plain}
\bibliography{references}

@misc{abadi_deep_2016,
  title = {Deep {{Learning}} with {{Differential Privacy}}},
  author = {Abadi, Mart{\'i}n and Chu, Andy and Goodfellow, Ian and McMahan, H. Brendan and Mironov, Ilya and Talwar, Kunal and Zhang, Li},
  year = {2016},
  month = jul,
  journal = {arXiv.org},
  eprint = {1607.00133},
  archiveprefix = {arXiv},
  urldate = {2025-06-11},
  howpublished = {https://arxiv.org/abs/1607.00133v2},
  langid = {english}
}

@article{adams_fidelity_2025,
  title = {On the Fidelity versus Privacy and Utility Trade-off of Synthetic Patient Data},
  author = {Adams, Tim and Birkenbihl, Colin and Otte, Karen and Ng, Hwei Geok and Rieling, Jonas Adrian and N{\"a}her, Anatol-Fiete and Sax, Ulrich and Prasser, Fabian and Fr{\"o}hlich, Holger},
  year = {2025},
  month = may,
  journal = {iScience},
  volume = {28},
  number = {5},
  pages = {112382},
  issn = {2589-0042},
  doi = {10.1016/j.isci.2025.112382},
  urldate = {2025-06-12}
}

@article{apellaniz_synthetic_2024,
  title = {Synthetic {{Tabular Data Validation}}: {{A Divergence-Based Approach}}},
  shorttitle = {Synthetic {{Tabular Data Validation}}},
  author = {Apell{\'a}niz, Patricia A. and Jim{\'e}nez, Ana and Arroyo Galende, Borja and Parras, Juan and Zazo, Santiago},
  year = {2024},
  journal = {IEEE Access},
  volume = {12},
  pages = {103895--103907},
  issn = {2169-3536},
  doi = {10.1109/ACCESS.2024.3434582},
  urldate = {2025-06-18}
}

@article{balasubramanian_isomap_2002,
  title = {The {{Isomap Algorithm}} and {{Topological Stability}}},
  author = {Balasubramanian, Mukund and Schwartz, Eric L.},
  year = {2002},
  month = jan,
  journal = {Science},
  volume = {295},
  number = {5552},
  pages = {7--7},
  publisher = {American Association for the Advancement of Science},
  doi = {10.1126/science.295.5552.7a},
  urldate = {2025-06-17}
}

@article{bejani_systematic_2021,
  title = {A Systematic Review on Overfitting Control in Shallow and Deep Neural Networks},
  author = {Bejani, Mohammad Mahdi and Ghatee, Mehdi},
  year = {2021},
  month = dec,
  journal = {Artif. Intell. Rev.},
  volume = {54},
  number = {8},
  pages = {6391--6438},
  issn = {0269-2821},
  doi = {10.1007/s10462-021-09975-1},
  urldate = {2025-06-11}
}

@article{bellman_mathematical_1959,
  title = {A Mathematical Theory of Adaptive Control Processes},
  author = {Bellman, Richard and Kalaba, Robert},
  year = {1959},
  month = aug,
  journal = {Proceedings of the National Academy of Sciences},
  volume = {45},
  number = {8},
  pages = {1288--1290},
  publisher = {Proceedings of the National Academy of Sciences},
  doi = {10.1073/pnas.45.8.1288},
  urldate = {2025-06-05}
}

@article{bengio_representation_2013,
  title = {Representation {{Learning}}: {{A Review}} and {{New Perspectives}}},
  shorttitle = {Representation {{Learning}}},
  author = {Bengio, Yoshua and Courville, Aaron and Vincent, Pascal},
  year = {2013},
  month = aug,
  journal = {IEEE Transactions on Pattern Analysis and Machine Intelligence},
  volume = {35},
  number = {8},
  pages = {1798--1828},
  issn = {1939-3539},
  doi = {10.1109/TPAMI.2013.50},
  urldate = {2025-06-12}
}

@article{bond-taylor_deep_2022,
  title = {Deep {{Generative Modelling}}: {{A Comparative Review}} of {{VAEs}}, {{GANs}}, {{Normalizing Flows}}, {{Energy-Based}} and {{Autoregressive Models}}},
  shorttitle = {Deep {{Generative Modelling}}},
  author = {{Bond-Taylor}, Sam and Leach, Adam and Long, Yang and Willcocks, Chris G.},
  year = {2022},
  month = nov,
  journal = {IEEE Transactions on Pattern Analysis and Machine Intelligence},
  volume = {44},
  number = {11},
  pages = {7327--7347},
  issn = {1939-3539},
  doi = {10.1109/TPAMI.2021.3116668},
  urldate = {2025-06-17}
}

@inproceedings{carlini_extracting_2021,
  title = {Extracting {{Training Data}} from {{Large Language Models}}},
  booktitle = {30th {{USENIX Security Symposium}} ({{USENIX Security}} 21)},
  author = {Carlini, Nicholas and Tram{\`e}r, Florian and Wallace, Eric and Jagielski, Matthew and {Herbert-Voss}, Ariel and Lee, Katherine and Roberts, Adam and Brown, Tom and Song, Dawn and Erlingsson, {\'U}lfar and Oprea, Alina and Raffel, Colin},
  year = {2021},
  month = aug,
  pages = {2633--2650},
  publisher = {USENIX Association},
  isbn = {978-1-939133-24-3}
}

@article{creswell_generative_2018,
  title = {Generative {{Adversarial Networks}}: {{An Overview}}},
  shorttitle = {Generative {{Adversarial Networks}}},
  author = {Creswell, Antonia and White, Tom and Dumoulin, Vincent and Arulkumaran, Kai and Sengupta, Biswa and Bharath, Anil A.},
  year = {2018},
  month = jan,
  journal = {IEEE Signal Processing Magazine},
  volume = {35},
  number = {1},
  pages = {53--65},
  issn = {1558-0792},
  doi = {10.1109/MSP.2017.2765202},
  urldate = {2025-06-12}
}

@article{damico_synthetic_2023,
  title = {Synthetic {{Data Generation}} by {{Artificial Intelligence}} to {{Accelerate Research}} and {{Precision Medicine}} in {{Hematology}}},
  author = {D'Amico, Saverio and Dall'Olio, Daniele and Sala, Claudia and Dall'Olio, Lorenzo and Sauta, Elisabetta and Zampini, Matteo and Asti, Gianluca and Lanino, Luca and Maggioni, Giulia and Campagna, Alessia and Ubezio, Marta and Russo, Antonio and Bicchieri, Maria Elena and Riva, Elena and Tentori, Cristina A. and Travaglino, Erica and Morandini, Pierandrea and Savevski, Victor and Santoro, Armando and {Prada-Luengo}, I{\~n}igo and Krogh, Anders and Santini, Valeria and Kordasti, Shahram and Platzbecker, Uwe and {Diez-Campelo}, Maria and Fenaux, Pierre and Haferlach, Torsten and Castellani, Gastone and Della Porta, Matteo Giovanni},
  year = {2023},
  month = jun,
  journal = {JCO Clinical Cancer Informatics},
  number = {7},
  pages = {e2300021},
  publisher = {Wolters Kluwer},
  doi = {10.1200/CCI.23.00021},
  urldate = {2025-06-12}
}

@article{deng_mnist_2012,
  title = {The Mnist Database of Handwritten Digit Images for Machine Learning Research},
  author = {Deng, Li},
  year = {2012},
  journal = {IEEE Signal Processing Magazine},
  volume = {29},
  number = {6},
  pages = {141--142},
  publisher = {IEEE},
  issn = {1053-587X},
  doi = {10.1109/MSP.2012.2205597},
  urldate = {2025-06-05}
}

@misc{directorate-general_for_health_and_food_safety_dg_sante_european_commission_regulation_2025,
  title = {Regulation ({{EU}}) 2025/327 of the {{European Parliament}} and of the {{Council}} on the {{European Health Data Space}}},
  author = {{Directorate-General for Health and Food Safety (DG SANTE), European Commission}},
  year = {2025},
  month = jan,
  howpublished = {https://eur-lex.europa.eu/legal-content/EN/TXT/?uri=CELEX%3A32025R0327},
  urldate = {2025-06-12}
}

@article{dockhorn_differentially_2023,
  title = {Differentially {{Private Diffusion Models}}},
  author = {Dockhorn, Tim and Cao, Tianshi and Vahdat, Arash and Kreis, Karsten},
  year = {2023},
  month = may,
  journal = {Transactions on Machine Learning Research},
  volume = {2},
  pages = {1--25},
  issn = {2835-8856},
  doi = {10.48550/arXiv.2302.05026},
  urldate = {2025-06-11}
}

@article{eraslan_deep_2019,
  title = {Deep Learning: New Computational Modelling Techniques for Genomics},
  shorttitle = {Deep Learning},
  author = {Eraslan, G{\"o}kcen and Avsec, {\v Z}iga and Gagneur, Julien and Theis, Fabian J.},
  year = {2019},
  month = jul,
  journal = {Nature Reviews Genetics},
  volume = {20},
  number = {7},
  pages = {389--403},
  publisher = {Nature Publishing Group},
  issn = {1471-0064},
  doi = {10.1038/s41576-019-0122-6},
  urldate = {2025-06-12},
  copyright = {2019 Springer Nature Limited},
  langid = {english}
}

@book{forsyth_computer_2002,
  title = {Computer Vision: A Modern Approach},
  author = {Forsyth, David A and Ponce, Jean},
  year = {2002},
  publisher = {prentice hall professional technical reference}
}

@article{galende_membership_2025,
  title = {Membership {{Inference Attacks}} and {{Differential Privacy}}: A Study within the Context of {{Generative Models}}},
  shorttitle = {Membership {{Inference Attacks}} and {{Differential Privacy}}},
  author = {Galende, Borja Arroyo and Apell{\'a}niz, Patricia A. and Parras, Juan and Zazo, Santiago and Uribe, Silvia},
  year = {2025},
  journal = {IEEE Open Journal of the Computer Society},
  pages = {1--10},
  issn = {2644-1268},
  doi = {10.1109/OJCS.2025.3572244},
  urldate = {2025-06-06}
}

@article{giuffre_harnessing_2023,
  title = {Harnessing the Power of Synthetic Data in Healthcare: Innovation, Application, and Privacy},
  shorttitle = {Harnessing the Power of Synthetic Data in Healthcare},
  author = {Giuffr{\`e}, Mauro and Shung, Dennis L.},
  year = {2023},
  month = oct,
  journal = {npj Digital Medicine},
  volume = {6},
  number = {1},
  pages = {1--8},
  publisher = {Nature Publishing Group},
  issn = {2398-6352},
  doi = {10.1038/s41746-023-00927-3},
  urldate = {2025-06-12},
  copyright = {2023 The Author(s)},
  langid = {english}
}

@article{gonzales_synthetic_2023,
  title = {Synthetic Data in Health Care: {{A}} Narrative Review},
  shorttitle = {Synthetic Data in Health Care},
  author = {Gonzales, Aldren and Guruswamy, Guruprabha and Smith, Scott R.},
  year = {2023},
  month = jan,
  journal = {PLOS Digital Health},
  volume = {2},
  number = {1},
  pages = {e0000082},
  issn = {2767-3170},
  doi = {10.1371/journal.pdig.0000082},
  urldate = {2025-06-12},
  pmcid = {PMC9931305},
  pmid = {36812604}
}

@book{goodfellow_deep_2016,
  title = {Deep {{Learning}}},
  author = {Goodfellow, Ian and Bengio, Yoshua and Courville, Aaron},
  year = {2016},
  month = oct,
  publisher = {The MIT Press},
  isbn = {978-0-262-03561-3}
}

@article{goodfellow_generative_2020,
  title = {Generative Adversarial Networks},
  author = {Goodfellow, Ian and {Pouget-Abadie}, Jean and Mirza, Mehdi and Xu, Bing and {Warde-Farley}, David and Ozair, Sherjil and Courville, Aaron and Bengio, Yoshua},
  year = {2020},
  month = oct,
  journal = {Commun. ACM},
  volume = {63},
  number = {11},
  pages = {139--144},
  issn = {0001-0782},
  doi = {10.1145/3422622},
  urldate = {2025-06-05}
}

@misc{guo_grey-box_2025,
  title = {A {{Grey-box Attack}} against {{Latent Diffusion Model-based Image Editing}} by {{Posterior Collapse}}},
  author = {Guo, Zhongliang and Lei, Chun Tong and Fang, Lei and Zhao, Shuai and Qian, Yifei and Lin, Jingyu and Wang, Zeyu and Chen, Cunjian and Arandjelovi{\'c}, Ognjen and Lau, Chun Pong},
  year = {2025},
  month = feb,
  number = {arXiv:2408.10901},
  eprint = {2408.10901},
  primaryclass = {cs},
  publisher = {arXiv},
  doi = {10.48550/arXiv.2408.10901},
  urldate = {2025-06-06},
  archiveprefix = {arXiv}
}

@inproceedings{ho_denoising_2020,
  title = {Denoising Diffusion Probabilistic Models},
  booktitle = {Proceedings of the 34th {{International Conference}} on {{Neural Information Processing Systems}}},
  author = {Ho, Jonathan and Jain, Ajay and Abbeel, Pieter},
  year = {2020},
  month = dec,
  series = {{{NIPS}} '20},
  pages = {6840--6851},
  publisher = {Curran Associates Inc.},
  address = {Red Hook, NY, USA},
  doi = {10.5555/2020.13903.34951},
  urldate = {2025-07-02},
  isbn = {978-1-71382-954-6}
}

@misc{jordon_synthetic_2024,
  title = {Synthetic {{Data}} -- What, Why and How?},
  author = {Jordon, James and Szpruch, Lukasz and Houssiau, Florimond and Bottarelli, Mirko and Cherubin, Giovanni and Maple, Carsten and Cohen, Samuel N. and Weller, Adrian},
  year = {2024},
  publisher = {The Royal Society},
  address = {London, UK}
}

@article{kaabachi_scoping_2025,
  title = {A Scoping Review of Privacy and Utility Metrics in Medical Synthetic Data},
  author = {Kaabachi, Bayrem and Despraz, J{\'e}r{\'e}mie and Meurers, Thierry and Otte, Karen and Halilovic, Mehmed and Kulynych, Bogdan and Prasser, Fabian and Raisaro, Jean Louis},
  year = {2025},
  month = jan,
  journal = {NPJ Digital Medicine},
  volume = {8},
  pages = {60},
  issn = {2398-6352},
  doi = {10.1038/s41746-024-01359-3},
  urldate = {2025-06-12},
  pmcid = {PMC11772694},
  pmid = {39870798}
}

@article{kanamori_theoretical_2010,
  title = {Theoretical {{Analysis}} of {{Density Ratio Estimation}}},
  author = {Kanamori, Takafumi and Suzuki, Taiji and Sugiyama, Masashi},
  year = {2010},
  journal = {IEICE Transactions on Fundamentals of Electronics, Communications and Computer Sciences},
  volume = {E93-A},
  number = {4},
  pages = {787--798},
  issn = {0916-8508, 1745-1337},
  doi = {10.1587/transfun.E93.A.787},
  urldate = {2025-06-09},
  langid = {english}
}

@inproceedings{kifer_no_2011,
  title = {No Free Lunch in Data Privacy},
  booktitle = {Proceedings of the 2011 {{ACM SIGMOD International Conference}} on {{Management}} of Data},
  author = {Kifer, Daniel and Machanavajjhala, Ashwin},
  year = {2011},
  month = jun,
  series = {{{SIGMOD}} '11},
  pages = {193--204},
  publisher = {Association for Computing Machinery},
  address = {New York, NY, USA},
  doi = {10.1145/1989323.1989345},
  urldate = {2025-07-22},
  isbn = {978-1-4503-0661-4}
}

@inproceedings{kingma_auto-encoding_2014,
  title = {Auto-{{Encoding Variational Bayes}}},
  booktitle = {Proceedings of the 2nd {{International Conference}} on {{Learning Representations}} ({{ICLR}})},
  author = {Kingma, Diederik P. and Welling, Max},
  year = {2014},
  eprint = {1401.4080},
  archiveprefix = {arXiv},
  doi = {10.48550/arXiv.1401.4080},
  urldate = {2025-06-05}
}

@article{lautrup_syntheval_2024,
  title = {Syntheval: A Framework for Detailed Utility and Privacy Evaluation of Tabular Synthetic Data},
  shorttitle = {Syntheval},
  author = {Lautrup, Anton D. and Hyrup, Tobias and Zimek, Arthur and {Schneider-Kamp}, Peter},
  year = {2024},
  month = dec,
  journal = {Data Mining and Knowledge Discovery},
  volume = {39},
  number = {1},
  pages = {1--6},
  issn = {1573-756X},
  doi = {10.1007/s10618-024-01081-4},
  urldate = {2025-06-18},
  langid = {english}
}

@incollection{lecun_convolutional_1998,
  title = {Convolutional Networks for Images, Speech, and Time Series},
  booktitle = {The Handbook of Brain Theory and Neural Networks},
  author = {LeCun, Yann and Bengio, Yoshua},
  year = {1998},
  month = oct,
  pages = {255--258},
  publisher = {MIT Press},
  address = {Cambridge, MA, USA},
  urldate = {2025-06-05},
  isbn = {978-0-262-51102-5}
}

@article{loaiza_ganem_diagnosing_2022,
  title = {Diagnosing and {{Fixing Manifold Overfitting}} in {{Deep Generative Models}}},
  author = {{Loaiza-Ganem}, Gabriel and Ross, Brendan Leigh and Cresswell, Jesse C. and Caterini, Anthony L.},
  year = {2022},
  month = jan,
  journal = {Transactions on Machine Learning Research},
  volume = {1},
  pages = {1--34},
  issn = {2835-8856},
  doi = {10.48550/arXiv.2201.02204},
  urldate = {2025-06-09}
}

@misc{lu_machine_2024,
  title = {Machine {{Learning}} for {{Synthetic Data Generation}}: {{A Review}}},
  shorttitle = {Machine {{Learning}} for {{Synthetic Data Generation}}},
  author = {Lu, Yingzhou and Shen, Minjie and Wang, Huazheng and Wang, Xiao and van Rechem, Capucine and Fu, Tianfan and Wei, Wenqi},
  year = {2024},
  month = jun,
  number = {arXiv:2302.04062},
  eprint = {2302.04062},
  primaryclass = {cs},
  publisher = {arXiv},
  doi = {10.48550/arXiv.2302.04062},
  urldate = {2025-06-17},
  archiveprefix = {arXiv}
}

@article{lv_which_2021,
  title = {Which {{GAN}}? {{A}} Comparative Study of Generative Adversarial Network-Based Fast {{MRI}} Reconstruction},
  shorttitle = {Which {{GAN}}?},
  author = {Lv, Jun and Zhu, Jin and Yang, Guang},
  year = {2021},
  month = may,
  journal = {Philosophical Transactions of the Royal Society A: Mathematical, Physical and Engineering Sciences},
  volume = {379},
  number = {2200},
  pages = {20200203},
  publisher = {Royal Society},
  doi = {10.1098/rsta.2020.0203},
  urldate = {2025-06-11}
}

@article{maaten_visualizing_2008,
  title = {Visualizing {{Data}} Using T-{{SNE}}},
  author = {van der Maaten, Laurens and Hinton, Geoffrey},
  year = {2008},
  journal = {Journal of Machine Learning Research},
  volume = {9},
  number = {86},
  pages = {2579--2605},
  issn = {1533-7928},
  doi = {10.48550/arXiv.0802.3284},
  urldate = {2025-06-17}
}

@article{mack_multivariate_1979,
  title = {Multivariate {\emph{K}}-Nearest Neighbor Density Estimates},
  author = {Mack, Y. P and Rosenblatt, M},
  year = {1979},
  month = mar,
  journal = {Journal of Multivariate Analysis},
  volume = {9},
  number = {1},
  pages = {1--15},
  issn = {0047-259X},
  doi = {10.1016/0047-259X(79)90065-4},
  urldate = {2025-07-02}
}

@article{mcinnes_umap_2018,
  title = {{{UMAP}}: {{Uniform Manifold Approximation}} and {{Projection}}},
  author = {McInnes, Leland and Healy, John and Saul, Nathaniel and Gro{\ss}berger, Lukas},
  year = {2018},
  journal = {Journal of Open Source Software},
  volume = {3},
  number = {29},
  pages = {861},
  publisher = {The Open Journal},
  doi = {10.21105/joss.00861}
}

@article{mendes_synthetic_2025,
  title = {Synthetic Data Generation: A Privacy-Preserving Approach to Accelerate Rare Disease Research},
  shorttitle = {Synthetic Data Generation},
  author = {Mendes, Jorge M. and Barbar, Aziz and Refaie, Marwa},
  year = {2025},
  month = mar,
  journal = {Frontiers in Digital Health},
  volume = {7},
  publisher = {Frontiers},
  issn = {2673-253X},
  doi = {10.3389/fdgth.2025.1563991},
  urldate = {2025-06-12},
  langid = {english}
}

@article{rashidi_novel_2024,
  title = {A Novel and Fully Automated Platform for Synthetic Tabular Data Generation and Validation},
  author = {Rashidi, Hooman H. and Albahra, Samer and Rubin, Brian P. and Hu, Bo},
  year = {2024},
  month = oct,
  journal = {Scientific Reports},
  volume = {14},
  number = {1},
  pages = {23312},
  publisher = {Nature Publishing Group},
  issn = {2045-2322},
  doi = {10.1038/s41598-024-73608-0},
  urldate = {2025-06-12},
  copyright = {2024 The Author(s)},
  langid = {english}
}

@inproceedings{rezende_variational_2015,
  title = {Variational {{Inference}} with {{Normalizing Flows}}},
  booktitle = {Proceedings of the 32nd {{International Conference}} on {{Machine Learning}}},
  author = {Rezende, Danilo and Mohamed, Shakir},
  year = {2015},
  month = jun,
  pages = {1530--1538},
  publisher = {PMLR},
  issn = {1938-7228},
  doi = {10.5555/2015.29170.1530},
  urldate = {2025-06-17},
  langid = {english}
}

@inproceedings{roelofs_meta-analysis_2019,
  title = {A {{Meta-Analysis}} of {{Overfitting}} in {{Machine Learning}}},
  booktitle = {Advances in {{Neural Information Processing Systems}}},
  author = {Roelofs, Rebecca and Shankar, Vaishaal and Recht, Benjamin and {Fridovich-Keil}, Sara and Hardt, Moritz and Miller, John and Schmidt, Ludwig},
  year = {2019},
  month = dec,
  pages = {14854--14864},
  volume = {32},
  publisher = {Curran Associates, Inc.},
  doi = {10.5555/2019.34680.14854},
  urldate = {2025-06-11}
}

@inproceedings{ronneberger_u-net_2015,
  title = {U-{{Net}}: {{Convolutional Networks}} for {{Biomedical Image Segmentation}}},
  shorttitle = {U-{{Net}}},
  booktitle = {Medical {{Image Computing}} and {{Computer-Assisted Intervention}} -- {{MICCAI}} 2015},
  author = {Ronneberger, Olaf and Fischer, Philipp and Brox, Thomas},
  editor = {Navab, Nassir and Hornegger, Joachim and Wells, William M. and Frangi, Alejandro F.},
  year = {2015},
  pages = {234--241},
  publisher = {Springer International Publishing},
  address = {Cham},
  doi = {10.1007/978-3-319-24574-4_28},
  isbn = {978-3-319-24574-4},
  langid = {english}
}

@article{ruthotto_introduction_2021,
  title = {An Introduction to Deep Generative Modeling},
  author = {Ruthotto, Lars and Haber, Eldad},
  year = {2021},
  journal = {GAMM-Mitteilungen},
  volume = {44},
  number = {2},
  pages = {e202100008},
  issn = {1522-2608},
  doi = {10.1002/gamm.202100008},
  urldate = {2025-06-05},
  copyright = {{\copyright} 2021 Wiley-VCH GmbH},
  langid = {english}
}

@inproceedings{selvaraju_grad-cam_2017,
  title = {Grad-{{CAM}}: {{Visual Explanations}} from {{Deep Networks}} via {{Gradient-Based Localization}}},
  shorttitle = {Grad-{{CAM}}},
  booktitle = {2017 {{IEEE International Conference}} on {{Computer Vision}} ({{ICCV}})},
  author = {Selvaraju, Ramprasaath R. and Cogswell, Michael and Das, Abhishek and Vedantam, Ramakrishna and Parikh, Devi and Batra, Dhruv},
  year = {2017},
  month = oct,
  pages = {618--626},
  issn = {2380-7504},
  doi = {10.1109/ICCV.2017.74},
  urldate = {2025-07-21}
}

@misc{sidorenko_benchmarking_2025,
  title = {Benchmarking {{Synthetic Tabular Data}}: {{A Multi-Dimensional Evaluation Framework}}},
  shorttitle = {Benchmarking {{Synthetic Tabular Data}}},
  author = {Sidorenko, Andrey and Platzer, Michael and Scriminaci, Mario and Tiwald, Paul},
  year = {2025},
  month = apr,
  number = {arXiv:2504.01908},
  eprint = {2504.01908},
  primaryclass = {cs},
  publisher = {arXiv},
  doi = {10.48550/arXiv.2504.01908},
  urldate = {2025-06-18},
  archiveprefix = {arXiv},
  journal = {arXiv.org}
}

@inproceedings{song_stochastic_2013,
  title = {Stochastic Gradient Descent with Differentially Private Updates},
  booktitle = {2013 {{IEEE Global Conference}} on {{Signal}} and {{Information Processing}}},
  author = {Song, Shuang and Chaudhuri, Kamalika and Sarwate, Anand D.},
  year = {2013},
  month = dec,
  pages = {245--248},
  doi = {10.1109/GlobalSIP.2013.6736861},
  urldate = {2025-06-11}
}

@misc{takahashi_differentially_2020,
  title = {Differentially {{Private Variational Autoencoders}} with {{Term-wise Gradient Aggregation}}},
  author = {Takahashi, Tsubasa and Takagi, Shun and Ono, Hajime and Komatsu, Tatsuya},
  year = {2020},
  month = jun,
  journal = {arXiv.org},
  eprint = {2006.11204},
  archiveprefix = {arXiv},
  doi = {10.48550/arXiv.2006.11204},
  urldate = {2025-06-11},
  howpublished = {https://arxiv.org/abs/2006.11204v1},
  langid = {english}
}

@article{takida_preventing_2022,
  title = {Preventing Oversmoothing in {{VAE}} via Generalized Variance Parameterization},
  author = {Takida, Yuhta and Liao, Wei-Hsiang and Lai, Chieh-Hsin and Uesaka, Toshimitsu and Takahashi, Shusuke and Mitsufuji, Yuki},
  year = {2022},
  month = oct,
  journal = {Neurocomputing},
  volume = {509},
  pages = {137--156},
  issn = {0925-2312},
  doi = {10.1016/j.neucom.2022.08.067},
  urldate = {2025-06-11}
}

@article{terrell_variable_1992,
  title = {Variable {{Kernel Density Estimation}}},
  author = {Terrell, George R. and Scott, David W.},
  year = {1992},
  journal = {The Annals of Statistics},
  volume = {20},
  number = {3},
  eprint = {2242011},
  eprinttype = {jstor},
  pages = {1236--1265},
  publisher = {Institute of Mathematical Statistics},
  issn = {0090-5364},
  urldate = {2025-06-05}
}

@article{tsirikoglou_survey_2020,
  title = {A {{Survey}} of {{Image Synthesis Methods}} for {{Visual Machine Learning}}},
  author = {Tsirikoglou, A. and Eilertsen, G. and Unger, J.},
  year = {2020},
  journal = {Computer Graphics Forum},
  volume = {39},
  number = {6},
  pages = {426--451},
  issn = {1467-8659},
  doi = {10.1111/cgf.14047},
  urldate = {2025-06-12},
  copyright = {{\copyright} 2020 The Authors. Computer Graphics Forum published by Eurographics - The European Association for Computer Graphics and John Wiley \& Sons Ltd},
  langid = {english}
}

@misc{us_department_of_health_and_human_services_office_for_civil_rights_notice_2025,
  title = {Notice of {{Proposed Rulemaking}} to {{Update}} the {{HIPAA Security Rule}}},
  author = {{U.S. Department of Health and Human Services, Office for Civil Rights}},
  year = {2025},
  month = jan,
  howpublished = {https://www.federalregister.gov/documents/2025/01/01/2025-00001/notice-of-proposed-rulemaking-to-update-the-hipaa-security-rule},
  urldate = {2025-06-12}
}

@article{walt_variable_2017,
  title = {Variable {{Kernel Density Estimation}} in {{High-Dimensional Feature Spaces}}},
  author = {van der Walt, Christiaan and Barnard, Etienne},
  year = {2017},
  month = feb,
  journal = {Proceedings of the AAAI Conference on Artificial Intelligence},
  volume = {31},
  number = {1},
  issn = {2374-3468},
  doi = {10.1609/aaai.v31i1.10885},
  urldate = {2025-06-05},
  copyright = {Copyright (c)},
  langid = {english}
}

@article{yang_medmnist_2023,
  title = {{{MedMNIST}} v2 - {{A}} Large-Scale Lightweight Benchmark for {{2D}} and {{3D}} Biomedical Image Classification},
  author = {Yang, Jiancheng and Shi, Rui and Wei, Donglai and Liu, Zequan and Zhao, Lin and Ke, Bilian and Pfister, Hanspeter and Ni, Bingbing},
  year = {2023},
  month = jan,
  journal = {Scientific Data},
  volume = {10},
  number = {1},
  pages = {41},
  publisher = {Nature Publishing Group},
  issn = {2052-4463},
  doi = {10.1038/s41597-022-01721-8},
  urldate = {2025-06-05},
  copyright = {2023 The Author(s)},
  langid = {english}
}

@article{zhang_no_2022,
  title = {No {{Free Lunch Theorem}} for {{Security}} and {{Utility}} in {{Federated Learning}}},
  author = {Zhang, Xiaojin and Gu, Hanlin and Fan, Lixin and Chen, Kai and Yang, Qiang},
  year = {2022},
  month = nov,
  journal = {ACM Trans. Intell. Syst. Technol.},
  volume = {14},
  number = {1},
  pages = {1:1--1:35},
  issn = {2157-6904},
  doi = {10.1145/3563219},
  urldate = {2025-07-22}
}

@article{zhu_generative_2024,
  title = {Generative {{Modeling}}},
  author = {Zhu, Jun-Yan and Isola, Phillip},
  year = {2024},
  month = jul,
  journal = {Open Encyclopedia of Cognitive Science},
  publisher = {MIT Press},
  doi = {10.21428/e2759450.ff10c6af},
  urldate = {2025-06-09},
  langid = {english}
}

\end{document}